\documentclass[letterpaper]{article} 
\usepackage{aaai24}  
\usepackage{times}  
\usepackage{helvet}  
\usepackage{courier}  
\usepackage[hyphens]{url}  
\usepackage{graphicx} 
\usepackage{natbib}  
\usepackage{caption} 
\usepackage{algorithm}
\usepackage{algorithmic}

\usepackage{newfloat}
\usepackage{listings}

\ifodd 0
\newcommand{\rev}[1]{\textcolor{blue}{#1}}
\newcommand{\revr}[1]{\textcolor{red}{#1}}

\newcommand{\com}[1]{\textbf{\color{red} (Comment: #1) }}
\newcommand{\comg}[1]{\textbf{\color{blue} (COMMENT: #1)}}
\newcommand{\response}[1]{\textbf{\color{blue} (RESPONSE: #1)}}
\newcommand{\deleting}[1]{\textcolor{brown}{#1}}
\else
\newcommand{\rev}[1]{#1}
\newcommand{\revr}[1]{#1}
\newcommand{\com}[1]{}
\newcommand{\comg}[1]{}
\newcommand{\response}[1]{}
\newcommand{\deleting}[1]{}
\fi

\usepackage{amsmath,amsfonts}
\usepackage{cancel}
\usepackage{graphicx}
\usepackage{textcomp}
\usepackage{dsfont}
\usepackage{xcolor}
\usepackage{bm}
\usepackage{amsmath}

\usepackage{etoolbox}
\usepackage{subfigure}
\usepackage{ifthen}
\newboolean{showcomments}
\setboolean{showcomments}{true}

\usepackage{verbatim}
\usepackage{booktabs}
\usepackage[mathscr]{euscript}
\usepackage{verbatim}
\usepackage{amssymb,amsmath,color,graphicx, amsbsy, bm}
\usepackage{epsfig}
\usepackage{amsthm}

\usepackage{fixltx2e}
\usepackage{xcolor}

\usepackage{amssymb}
\usepackage{algorithmic}       
\usepackage{amsmath}
\usepackage{graphicx}
\usepackage{booktabs}
\usepackage[flushleft]{threeparttable}
\usepackage{multirow}
\usepackage{amsthm}
\usepackage{graphicx,booktabs,multirow}
\usepackage{stmaryrd}
\usepackage{multirow,url}
\usepackage{algorithm}
\usepackage{color}

\usepackage{mathtools}

\newcommand{\bs}{\boldsymbol}

\newtheorem*{kquestion}{Key Question}
\newtheorem{theorem}{Theorem}

\newtheorem{proposition}{Proposition}
\newtheorem{lemma}{Lemma}

\newtheorem{definition}{Definition}

\def\T{\mathcal{T}}
\def\K{\mathcal{K}}
\def\I{\mathcal{I}}
\def\N{\mathcal{N}}
\def\M{\mathcal{M}}

\def\s{\boldsymbol{s}}
\def\a{\boldsymbol{a}}

\def\As{\mathcal{A}}

\def\thetae{\boldsymbol{\theta}}

\def\Atmp{\boldsymbol{a}^{\text{Next}}}

\def\task{k}

\def\N{\mathcal{N}}
\def\M{\mathcal{M}}
\def\K{\mathcal{K}}

\def\taul{\tau^{\rm {local}}}
\def\taut{\tau^{\rm {tran}}}
\def\taue{\tau^{\rm {edge}}}

\def\we{w^{\rm {edge}}}

\newcommand\norm[1]{\lVert#1\rVert}

\def\Su{\boldsymbol{s}^{\textsc{u}}}
\def\su{\boldsymbol{s}^{\textsc{u}}}
\def\so{\boldsymbol{s}^{\textsc{o}}}
\def\au{\boldsymbol{a}^{\textsc{u}}}
\def\ah{\boldsymbol{a}^{\textsc{h}}}
\def\ao{\boldsymbol{a}^{\textsc{o}}}
\def\piu{{\pi}^{\textsc{u}}}
\def\pio{{\pi}^{\textsc{o}}}

\def\Su{\mathcal{S}^{\textsc{u}}}
\def\So{\mathcal{S}^{\textsc{o}}}
\def\Ag{\mathcal{A}^{\textsc{u}}}
\def\Ao{\mathcal{A}^{\textsc{o}}}
\def\Do{\mathcal{D}^{\textsc{o}}}
\def\Du{\mathcal{D}^{\textsc{u}}}

\def\q{\boldsymbol{q}}
\def\pii{\boldsymbol{\pi}}

\def\thetao{\boldsymbol{\theta}^{\textsc{o}}}
\def\thetaom{\boldsymbol{\theta}^{\textsc{o}-}}
\def\thetamu{\boldsymbol{\theta}^{\mu}}
\def\thetamum{\boldsymbol{\theta}^{\mu-}}
\def\thetau{\boldsymbol{\theta}^{\textsc{u}}}
\def\thetaum{\boldsymbol{\theta}^{\textsc{u}-}}
\def\qo{Q^{\textsc{o}}}
\def\qu{Q^{\textsc{u}}}
\def\qt{Q^{\textsc{trg}}}

\def\nete{\text{Net}\_{\text{Eval}}}
\def\nett{\text{Net}\_{\text{TRG}}}

\def\neta{\text{Net}\_{\text{Act}}}
\def\netc{\text{Net}\_{\text{Crt}}}
\def\netat{\text{Net}\_{\text{ActTRG}}}
\def\netct{\text{Net}\_{\text{CrtTRG}}}

\DeclareCaptionStyle{ruled}{labelfont=normalfont,labelsep=colon,strut=off} 
\floatstyle{ruled}
\newfloat{listing}{tb}{lst}{}
\floatname{listing}{Listing}

\title{Fractional Deep Reinforcement Learning for \\ Age-Minimal  Mobile Edge Computing}
\author{
    Lyudong Jin\equalcontrib\textsuperscript{\rm 1},
    Ming Tang\equalcontrib\textsuperscript{\rm 2},
    Meng Zhang\thanks{Corresponding author.}\textsuperscript{\rm 1},
    Hao Wang \textsuperscript{\rm 3}
}
\affiliations{
    \textsuperscript{\rm 1}Zhejiang University, Haining, Zhejiang China\\    
    \textsuperscript{\rm 2}Southern University of Science and Technology, Shenzhen, Guangdong China\\
    \textsuperscript{\rm 3}Monash University, Melbourne, Victoria Australia\\
     lvdong.22@intl.zju.edu.cn,tangm3@sustech.edu.cn, mengzhang@intl.zju.edu.cn, hao.wang2@monash.edu
    
}

\begin{document}

\maketitle

\begin{abstract}
Mobile edge computing (MEC) is a promising paradigm for real-time applications with intensive computational needs (e.g., autonomous driving), as it can reduce the processing delay.
{In this work, we focus on the timeliness of computational-intensive updates, measured by \textit{Age-of-Information (AoI)}, and study how to jointly optimize the task updating and offloading  policies for AoI with fractional form.
Specifically, we consider edge load dynamics and formulate a task scheduling problem to minimize the expected time-average AoI.
The uncertain edge load dynamics, the nature of the fractional objective, and hybrid continuous-discrete action space {(due to the joint optimization)} make this problem challenging and existing approaches not directly applicable. 
To this end, we propose a fractional reinforcement learning (RL) framework 
and prove its convergence. 
We further design a model-free fractional deep RL (DRL) algorithm, where each device makes scheduling decisions with the hybrid action space without knowing the system dynamics and decisions of other devices. 
\revr{Experimental results show that our proposed algorithms reduce the average AoI {by up to $57.6\%$}}
 compared with several non-fractional benchmarks.} 
\end{abstract}

\section{Introduction}

\subsection{Background and Motivations}
The next-generation network demands mobile devices (e.g., smartphones and Internet-of-Things devices) to generate zillions of bytes of data and  accomplish unprecedentedly computationally intensive tasks. 
Mobile devices, however, will be unable to timely process all their tasks locally due to their limited computational resources. 
To fulfill the low latency requirement, mobile edge computing \cite{mao2017survey} (MEC), also known as multi-access edge computing \cite{porambage2018survey}, has  become an emerging paradigm distributing computational tasks and services from the network core to the network edge.
By enabling mobile devices to offload their computational tasks to nearby edge nodes, MEC can reduce the task processing delay.

On the other hand, the proliferation of real-time and computation-intensive applications (e.g., cyber-physical systems) has significantly boosted the demand for information freshness , e.g., \cite{AoISurvey2021,kaul2012real,shesher2022how}, in addition to low latency.
For example, the real-time velocity and location knowledge of the surrounding vehicles is crucial in achieving safe and efficient autonomous driving. Another emerging example is metaverse applications, in which users anticipate real-time  virtual reality services and real-time control over their avatars.
{In these applications, the experience of users depends on how fresh the received information is rather than how long it takes to receive that information.} Such a requirement motivates a new network performance metric, namely \textit{Age of Information (AoI)} \cite{AoISurvey2021,kaul2012real,shesher2022how}. It measures the time elapsed since the most up-to-date data (computational results) was received.


While the majority of existing studies on MEC were concerned about delay reduction \rev{(e.g., \cite{wang2021delay,Tang2020TMC}), most of real-time applications mentioned above concern about fresh status updates, while delay itself does not directly reflect timeliness.} Here we highlight the huge difference between delay and AoI. 
{Specifically, task delay takes into account only the duration between when the task is generated and when the task output has been received by the mobile device. 
Thus, under less frequent updates (i.e., when tasks are generated in a lower frequency), task delays are naturally smaller. 
This is because infrequent updates lead to empty queues and hence reduced queuing delays of the tasks. In contrast, AoI takes into account both the task delay and the freshness of the task output.  Thus, to minimize the AoI \rev{with computational-intensive tasks}, the update frequency needs to be neither too high nor too low in order to reduce the delay of each task while ensuring the freshness of the most up-to-date task output.}
More importantly, such a difference between delay and AoI leads to a counter-intuitive important phenomenon {in designing age-minimal scheduling policy:  upon the reception of each update, the mobile device  may need to \textit{wait} for a certain amount of time  to generate the next new task \cite{sun2017tit}.}

%

Therefore, the age-minimal MEC systems necessitate meticulous design of a scheduling policy for each mobile device, which should encompass two fundamental decisions.
The first decision is the \textit{updating} decision, i.e., upon completion of a task, how long should a mobile device wait for generating the next one. The second is the task  \textit{offloading} decision, i.e., whether to offload the task or not? If yes, which edge node to choose? \revr{Although existing  works on MEC have addressed the task offloading decision (e.g., \cite{Tang2020TMC,ma2022green,zhao2022multi,zhu2022federated,he2022age,chen2022info})} and \rev{some studies considered AoI (e.g., \cite{zhu2022federated,he2022age,chen2022info})}, they did not consider designing the task updating policy to improve data timeliness.

What's more, the age-minimal technique can also be extended to other ratio optimization problems including financial portfolio optimization \cite{keating2002universal} and energy efficiency (EE) maximization problems for wireless communications \cite{9709888}.

In this paper, we aim to answer the following question:
\begin{kquestion}
\rev{How should mobile devices optimize their updating and offloading policies of hybrid action space in dynamic MEC systems in order to minimize their fractional objectives of AoI?}
\end{kquestion}

\subsection{Solution Approach and Contributions}

In this work, we take into account system dynamics in MEC systems and aim at designing distributed AoI-minimal DRL algorithms to jointly optimize task updating and offloading. 
We first propose a novel fractional RL framework, incorporating reinforcement learning techniques and Dinkelbach's approach (for fractional programming) in \cite{dinkelbach67ms}. We further propose a fractional Q-Learning algorithm and 
analyze its convergence.
To  address the hybrid action space,  we further design a fractional DRL-based algorithm.
Our main contributions are summarized as follows:
\begin{itemize}
    \item \emph{Joint Task Updating and Offloading Problem:} We formulate the joint task updating and offloading problem that takes into account unknown system dynamics. \rev{\textit{To the best of our knowledge, this is the first work  designing the joint updating and offloading policy for age-minimal  MEC.}}
    \item \emph{Fractional RL Framework:} 
    To overcome fractional objective of the average AoI, we propose a novel fractional RL framework. We further propose a fractional Q-Learning algorithm. We design a stopping condition, leading to a provable linear convergence rate without the need of increasing inner-loop steps. 
    \item \emph{Fractional DRL Algorithm:} \rev{We overcome unknown dynamics and hybrid action space of offloading and updating decisions and propose a
    \revr{fractional DRL-based distributed scheduling algorithm for age-minimal MEC,} }
	\rev{\revr {which} extends the dueling double deep Q-network (D3QN) and deep deterministic policy gradient (DDPG) techniques  into our proposed fractional RL framework.}
    \item \emph{Performance Evaluation:} \revr {Our algorithm  significantly outperforms the benchmarks that neglect the fractional nature with an average AoI reduction by up to \revr{$57.6\%$}. In addition, the joint optimization of offloading and updating can further reduce the AoI \revr{by up to $31.3\%$.}}
\end{itemize}


\section{Literature Review}\label{sec:review}

\textbf{Mobile Edge Computing}:
Existing excellent works have conducted various research questions in MEC, including resource allocation (e.g., \cite{wang2022utility}), service placement (e.g., \cite{taka2022service}), and  proactive caching (e.g., \cite{liu2022distributed}). Task offloading \cite{wang2022decentralized,ma2022green,Chen_Xie_2022}, as another main research question in MEC, attracting considerable attention. To address the unknown system dynamics and reduce task delay, many existing works have proposed DRL-based approaches to optimize the task offloading in a centralized manner (e.g., \cite{huang2020deep,tuli2022dynamic}). \rev{As in our work, some existing works have proposed distributed DRL-based algorithms (e.g., \revr{\cite{Tang2020TMC,liu2022deep,zhao2022multi}}) which do not require the global information.}  
\emph{Despite the success of these works in reducing the task delay, these approaches are NOT easily applicable to age-minimal MEC due to the aforementioned challenges of  fractional objective and  hybrid action space.}

\textbf{Age of Information}: Kaul \textit{et al.} first  introduced AoI in \cite{kaul2012real}. Assuming complete and known statistical information, the majority of this line of work mainly focused on the optimization and analysis of AoI in queueing systems and wireless networks (see 
\revr{\cite{AoIMEC3,AoIMEC1,AoIMEC4}}, and a survey in \cite{AoISurvey2021}).
Zou \textit{et al.} in \cite{AoIMEC3},
Zhou \textit{et al.} in \cite{zhou2024}, Chiariotti \textit{et al.} in \cite{AoIMEC1}, and Kuang \textit{et al.} in \cite{AoIMEC4}.
  \emph{The above studies analyzed simple single-device-single-server models and hence did not consider offloading.}  

A few studies investigated DRL algorithm design to minimize AoI in various application scenarios, including wireless networks (e.g., \cite{ceran2021reinforcement}),  Internet-of-Things (e.g., \cite{AoIDeep2,wang2022distributed}),  vehicular networks (e.g., \cite{AoIDeep1}), and UAV-aided networks (e.g., \cite{UAVDeep1,UAVDeep2}). 
\emph{This line of work mainly focused on optimal resource allocation and trajectory design.} \deleting{A few \textbf{closely related studies} \cite{zhu2022federated,he2022age,chen2022info,xie2022reinforcement,xu2022aoi,zhu2022online}:}
{Existing works  
considered AoI as the performance metric for task offloading in MEC and proposed DRL-based approaches to address the AoI minimization problem.} 
Chen \emph{et al.} in \cite{chen2022info} considered AoI to capture the freshness of computation outcomes and proposed a multi-agent DRL algorithm. 
\emph{However, these works focused on designing task offloading policy but did not optimize updating policy.}
\textit{Most importantly,  all aforementioned approaches 
did not account for fractional RL and hence cannot directly tackle our considered problem.}

\textbf{RL with Fractional Objectives:} Research on RL with fractional objectives is currently limited.  Ren \textit{et. al.} introduced fractional MDP \cite{1431043}. Reference \cite{doi:10.1080/02522667.2015.1105525} further studied partially observed MDPs with fractional rewards. However, RL methods were not considered in these studies. Suttle \textit{et al.} \cite{pmlr-v139-suttle21a} proposed a two-timescale RL algorithm for the fractional cost, but it requires additional fixed reference states in the Q-learning update process to approximate the outer loop update and leave finite-time convergence analysis unsettled.


\section{System Model}\label{sec:model}

\begin{figure}
    \centering
    \includegraphics[height=2.3cm]{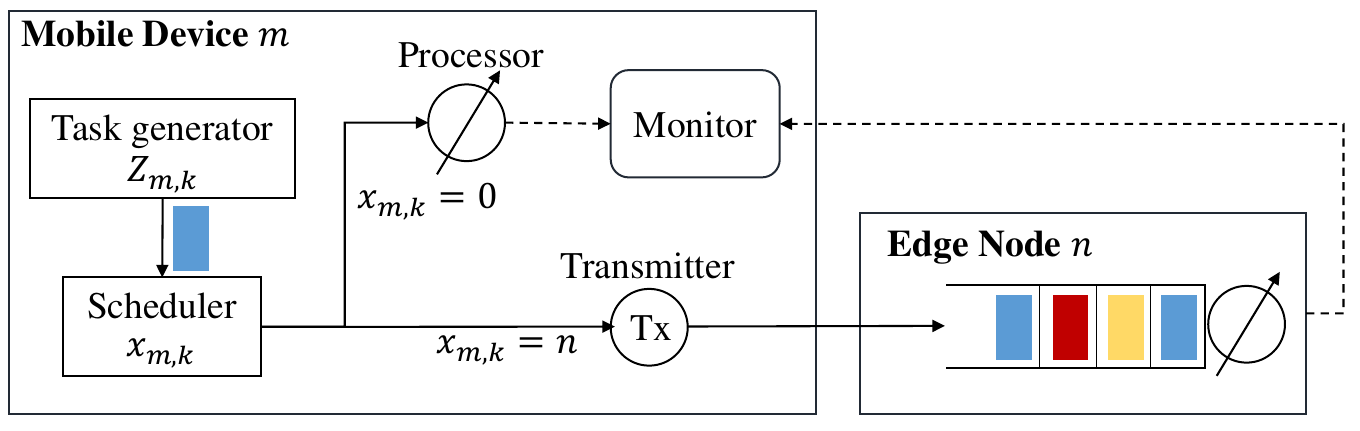}
    \caption{An illustration of an MEC system with a mobile device $m\in\M$ and an edge node $n\in\N$ \revr{where}the tasks offloaded by different mobile devices are represented using different colors.
    }
    \label{fig:model}
\end{figure}

Consider $M$ mobile devices and $N$ edge nodes, which are in set $\M=\{1,2,\ldots, M\}$ and set $\N=\{1,2,\ldots N\}$, respectively. \revr{ We consider an infinite-horizon continuous-time system model illustrated in Fig. \ref{fig:model}}\deleting{ with the system starting at time $t=0$}. \deleting{Mobile devices generate (computational) tasks over time and can select to process tasks locally or offload them to edge nodes (servers).}
\deleting{We present an illustrative system model in Fig. \ref{fig:model}.}
\subsection{Device Model}
Each mobile device $m\in\M$ has a task generator, a scheduler, and a monitor. The task generator generates new tasks for processing while the scheduler determines where to process the tasks\deleting{, i.e., process locally or offload to a specific edge node for processing.}\revr{ and the}  task output is sent to the monitor. We refer to a task output received by the monitor as an \textit{update}.

\textbf{Task Generator}: We consider a \textit{generate-at-will} model, as in \cite{sun2017tit}\deleting{\cite{arafa2019timely,Zhang21JSAC}}, i.e.,
each task generator can decide when to generate the next new task. At the time when task $k-1$ of mobile device $m\in\M$ has been completed (denoted by time $t'_{m,k-1}$), the task generator observes the task delay $Y_{m,k-1}$ and makes a decision on $Z_{m,k}$, i.e., the waiting time for generating the \emph{next} task $k$. Let $\su_m(k)$ and $\au_m(k)$ denote the state and action \revr{with task $k$ of  mobile device $m$ respectively:} 
   $ \su_m(k) = Y_{m,k-1}, ~\au_m(k) = \rev{Z_{m,k}}, ~  k\in\K, m\in\M$.

Let $\Su = (0,\bar{Y}]$ and \rev{$\Ag \in[0, \bar{Z}]$} denote the state and action space, respectively. Let $\piu_m: \Su\rightarrow\Ag$ denote the updating policy of mobile device $m\in\M$ that maps from $\Su$ to $\Ag$. Specifically,  let $t_{m,k}$ denote the time stamp when the task generator of mobile device $m$ generates task $k$, after which the task is sent to the scheduler. The transmission time is considered negligible and $t_{m,k+1} = t'_{m,k} + Z_{m,k+1}$. From \cite{sun2017tit}, the optimal waiting strategy may outperform the zero-wait policy, i.e., $Z_{m,k}$ may not necessarily be zero and requires proper optimization.

\deleting{\textbf{Task Generator}: We consider a \textit{generate-at-will} model, as in \cite{sun2017tit}\deleting{\cite{arafa2019timely,Zhang21JSAC}}, i.e.,
each task generator can decide when to generate the next new task. Specifically,  {let $t_{m,k}$ denote the time stamp when the task generator of mobile device $m$ generates task $k$, after which the task is sent to the scheduler. The transmission time is considered negligible.
After task $k$ is completed (i.e., when the output of task  $k$ has been received by monitor), the task generator decides when to generate the next task $k+1$. We use $Z_{m,k+1}\in(0,\bar{Z}]$ to denote the decision variable on the amount of time to wait (after completing task $k$) for generating the next task $k+1$, where $\bar{Z}$ is  a predefined value and is the maximum waiting time that the generator can choose. In other words, let $t_{m,k}'$ denote the time stamp when task $k$ is completed; then, $t_{m,k+1} = t'_{m,k} + Z_{m,k+1}$.  Based on \cite{sun2017tit}, the optimal age-minimal waiting strategy may outperform the zero-wait policy, i.e., decision variable $Z_{m,k}$ may not necessarily be zero and requires proper optimization.} }

\deleting{\subsubsection{Task Offloading Process}
At the time when task $k$ of mobile device $m\in\M$ is generated (i.e., time $t_{m,k}$), the task scheduler observes the queue lengths of edge nodes and makes the offloading decision. Let $\so_m(k)$ denote the state vector associated with task $k$ of mobile device $m$.
\begin{equation}
    \so_m(k) = \q(t_{m,k}), ~ k\in\K, m\in\M,
\end{equation}
where $\q(t_{m,k}) = (q_n(t_{m,k}), n\in\N)$ denotes the queue lengths of all edge nodes. We assume that edge nodes send their queue length information upon the requests of mobile devices. Since a generator generates a new task only after the previous task has been processed, the queue length  is less than or equal to $M$. Thus, it can be encoded in  $O (\log_2 M)$ bits, which incurs only small signaling overheads. Let $\So = \M^{1\times N}$ denote the state space. 

Let $\ao_m(k)$ denote the action associated with task $k$ of mobile device $m$. Thus, 
\begin{equation}
    \ao_m(k) = x_{m,k}, ~~k\in\K, m\in\M.
\end{equation}
Let $\Ao \in\{0\}\cup \N$ denote the offloading action space. Let $\pio_m$ denote the task offloading policy of mobile device $m\in\M$, which is a  mapping from $\So$ to $\Ao$. }

\textbf{Scheduler}: 
At the time when task $k$ of mobile device $m\in\M$ is generated (i.e., time $t_{m,k}$), the task scheduler observes the queue lengths of edge nodes and makes the offloading decision denoted by $x_{m,k}\in\{0\}\cup \N $. Let $\so_m(k)$ denote the state vector associated with task $k$ of mobile device $m$:
$\so_m(k) = \q(t_{m,k}), ~ k\in\K, m\in\M$, where $\q(t_{m,k}) = (q_n(t_{m,k}), n\in\N)$ corresponds to the queue lengths of all edge nodes. We assume that edge nodes send their queue length information upon the requests of mobile devices. Since a generator generates a new task only after the previous task has been processed, the queue length  is less than or equal to $M$. Thus, it can be encoded in  $O (\log_2 M)$ bits, which incurs only small signaling overheads. Let $\So = \M^{1\times N}$ denote the state space. Let $\ao_m(k)$ denote the action associated with task $k$ of mobile device $m$. Thus, $\ao_m(k) = x_{m,k}, ~~k\in\K, m\in\M$. Let $\Ao \in\{0\}\cup \N$ denote the offloading action space. Let $\pio_m$ denote the task offloading policy of mobile device $m\in\M$.
 \deleting{ If $x_{m,k}=0$, then task $k$ is processed locally. If $x_{m,k}=n\in\N$, then task $k$ is  sent to edge node $n$ for processing. {Note that since a new task can be generated only after the previously generated task has been completed, the mobile device does not need to maintain a queue for either the tasks to be processed locally or the tasks to be sent to edge nodes.} }


If mobile device $m$ processes task $k$ locally, 
then let $\taul_{m,k}$ (in seconds) denote the service time  of mobile device $m\in\M$ for processing task $k$. The value of $\taul_{m,k}$ depends on the size of task $k$ and the real-time processing capacity of mobile device $m$ (e.g., whether the device is busy in processing tasks of other applications), which are unknown \emph{a priori}. 
If mobile device $m$  offloads task $k$ to edge node $n\in\N$, then let 
$\taut_{n,m,k}$ (in seconds) denote the service time  of mobile device $m\in\M$ for sending task $k$ to edge node $n$. \rev{The value of $\taut_{n,m,k}$ depends on the time-varying wireless channels and  is unknown \emph{a priori}. We assume that   $\taul_{m,k}$ and $\taut_{n,m,k}$ are random variables that follow exponential distribution \cite{tang2021age,liu2022deep,zhu2022online}, respectively, which are independent of edge loads}.

\subsection{Edge Node Model}

Upon receiving a task offloaded by a mobile device, edge node $n\in\N$ enqueues the task for processing.  The queue may store the tasks offloaded by multiple mobile devices, as shown in Fig. \ref{fig:model}. \rev{Suppose the queue  operates in a first-in-first-out (FIFO) manner \cite{AoISurvey2021}.} 
 
Let  $\we_{n,m,k}$ (in seconds) denote the time duration that task $k$ of mobile device $m\in\M$ waits at the queue of edge node $n$. Let $\taue_{n, m,k}$ (in seconds) denote the service time  of edge node $n$ for processing task $k$ of mobile device $m$. \rev{The value of $\taue_{n,m,k}$ depends on the size of task $k$ and may be unknown \emph{a priori}. We assume that $\taue_{n,m,k}$ is a random variable  following exponential distribution \revr{as well}\deleting{\cite{tang2021age,liu2022deep,zhu2022online}}. In addition, the value of $\we_{n,m,k}$ depends on the processing time of the tasks placed in the queue (of edge node $n$) ahead of the task $k$ of device $m$, where those tasks are possibly offloaded by  mobile devices other than device $m$. Thus, since mobile device $m$ does not know the offloading behaviors of other mobile devices \emph{a priori}, it does not know the value of $\we_{n,m,k}$ beforehand.
}

\subsection{Age of Information}

The \textit{age of information (AoI)} for mobile device $m$ \rev{at time stamp $t$} \cite{AoISurvey2021} is given by 
\begin{align}
    \Delta_m(t)= t- U_{m}(t), ~~\forall m\in\mathcal{M}, t\geq 0,
\end{align}
where $U_{m}(t)\triangleq \max_{k} [t_{m,k}| t'_{m,k}\leq t]$ stands for the time stamp of the most recently completed task.

\rev{We use $Y_{m,k}\triangleq t_{m,k}'-t_{m,k}$ to denote the delay of task $k$, i.e., the time it takes to complete task $k$.} 
Thus, 
\begin{align}
  \hspace{-0.35cm}  Y_{m,k}=\begin{cases}
     \taul_{m,k},&x_{m,k}=0,\\
     \taut_{n,m,k} + \we_{n,m,k} + \taut_{n,m,k},&x_{m,k}=n\in\mathcal{N}.
    \end{cases}
\end{align}
\rev{We consider a \textit{drop time} $\bar{Y}$ (in seconds). That is, we assume that if a task has not been completely processed within $\bar{Y}$ seconds, the task will be dropped \cite{Tang2020TMC,li2020age}.} \rev{Meantime, the AoI keeps increasing until the next task is completed.}

To capture the overall performance of mobile device $m$, we define the trapezoid area associated with time interval $[t_{m,k},t_{m,k+1})$ \cite{AoISurvey2021}: 
\begin{align}
    A(Y_{m,k},Z_{m,k+1},Y_{m,k+1})\triangleq ~&\frac{1}{2}(Y_{m,k}+Z_{m,k+1}+Y_{m,k+1})^2\nonumber\\
    &-\frac{1}{2}Y_{m,k+1}^2. \label{trapezoid}
\end{align}
Based on \eqref{trapezoid}, we can characterize the objective of mobile device $m$, i.e., to minimize the time-average AoI of each device $m\in\mathcal{M}$: \cite{AoISurvey2021}
\begin{align}
    \revr{{\Delta}^{(ave)}_m\triangleq}&\liminf_{M\rightarrow \infty}\frac{ \sum_{m=1}^M A(Y_{m,k},Z_{m,k+1},Y_{m,k+1})}{ \sum_{m=1}^M (\revr{Y_{m,k}}+Z_{m,k+1})}.
\end{align}

\subsection{Problem Formulation}\label{subsec:prob}
Let $\pii_m=(\piu_m, \pio_m)$ denote the policy of mobile device $m\in\M$. This is a stationary policy that contains the mapping from $\Su\times \So$ to $\Ag\times \Ao$. 
Given a stationary policy $\pii_{m}$, the expected time-average AoI of mobile device $m\in\M$ is
\begin{align}
\mathbb{E}[\Delta_m^{(ave)}|\pii_m]\triangleq \frac{ \mathbb{E}[A(Y_{m,k},Z_{m,k+1},Y_{m,k+1})|\pii_m]}{ \mathbb{E}[\revr{Y_{m,k}}+Z_{m,k+1}|\pii_m]}.\label{Problem-fractional}
\end{align}
We take the expectation $\mathbb{E}[\cdot]$ over policy $\bs\pi_m$ and the time-varying system parameters, e.g., \revr{the time-varying processing duration as well as the edge load dynamics.}

 We \revr{aim at} the optimal policy $\pii_{m}^*$ for each mobile device $m\in\mathcal{M}$ to minimize its expected time-average AoI. 
 \begin{equation}\label{eq:opt-rl}
 \begin{aligned}
 \pii_{m}^* = \arg\mathop{\text{minimize}}_{\pii_{m}} \quad & \mathbb{E}\left[\Delta_m^{(ave)}|\pii_m\right]. 
 \end{aligned}
 \end{equation}
The fractional objective in \eqref{Problem-fractional} introduces a major challenge in designing the optimal policy, which is significantly different from conventional RL and DRL algorithms. Specifically,
the difficulty of directly expressing the immediate reward (cost) of each action for the fractional RL problem. \rev{Specifically, it seems to be  straightforward to define the reward  (or cost) function as either the instant AoI (i.e., $\Delta_m(t_{m,k}')$) \cite{chen2022info,he2022age} or the average AoI during certain time interval (e.g., $[t_{m,k},t_{m,k+1})$). However, consider the time-average over infinite time horizon,  neither  minimizing $\mathbb{E}[\Delta_m(t_{m,k}')|\pii_{m}]$ nor $\mathbb{E}[A(Y_{m,k},Z_{m,k+1},Y_{m,k+1})/(\revr{Y_{m,k}}+Z_{m,k+1})|\pii_{m}]$ is equivalent to minimizing \eqref{Problem-fractional}.} 




\section{Fractional RL Framework}\label{sec:solution}
In this section, we propose a fractional RL framework for solving Problem \eqref{eq:opt-rl}.\revr{We first} present a two-step reformulation of Problem \eqref{eq:opt-rl}. We then introduce the fractional RL framework, under which we present a fractional Q-Learning algorithm with provable convergence guarantees.

\subsubsection{Dinkelbach's Reformulation}
\revr{With the proposed Problem \ref{eq:opt-rl} we consider the Dinkelbach's reformulation and a discounted reformulation in the following.}
we define a reformulated AoI in an average-cost fashion:
\begin{align}
\mathbb{E}[\Delta_m'|\pii_m,\gamma] \!\triangleq &\!\! \lim_{K\rightarrow\infty}\!\frac{1}{K}\!\sum_{k=1}^{K}\!\left\{ \mathbb{E}[A(Y_{m,k},Z_{m,k+1},Y_{m,k+1})|\pii_m] \right. \nonumber\\
&\left. - \gamma \mathbb{E}[\revr{Y_{m,k}}+Z_{m,k+1}|\pii_m] \right\}. \label{eq:average}
\end{align}
 Let $\gamma^*$ be the optimal value of  Problem \eqref{eq:opt-rl}. Leveraging Dinkelbach's method \cite{dinkelbach67ms}, we have the following reformulated problem:
 \begin{lemma}[\cite{dinkelbach67ms}]\label{L1}
 Problem \eqref{eq:opt-rl} is equivalent to the following reformulated problem:
 \begin{align}
  \pii_m^*  = \arg\mathop{\rm{minimize}}_{\pii_{m}}~ \mathbb{E}[\Delta_m'|\pii_m,\gamma^*],\quad \forall m\in\mathcal{M},\label{eq:average-p}
 \end{align}
 where $\pii_m^*$ is the optimal solution to Problem \eqref{eq:opt-rl}.
 \end{lemma}
 
\rev{ Since $\mathbb{E}[\Delta_m'|\pii_m,\gamma^*]\geq 0$ for any $\pi$ and $\mathbb{E}[\Delta_m'|\pii_m^*,\gamma^*]= 0$, $\pii_m^*$ is also optimal to the Dinkelbach reformulation. This implies the reformulation equivalence is also established for our stationary policy space.}

\subsubsection{Discounted Reformulation} 
Following Dinkelbach's reformulation, we reformulate the problem in \eqref{eq:average-p} one step further by considering a discounted objective.
Let $\delta\in(0,1]$ be the discount factor, capturing how the objective is discounted in the future. We define 
\begin{align}
\mathbb{E}[\Delta_m^\delta|\pii_m,\gamma] \triangleq &\sum_{k=1}^{\infty} \delta^{k}\left\{\mathbb{E}[A(Y_{m,k},Z_{m,k+1},Y_{m,k+1})|\pii_m] \right. \nonumber\\
&\left. - \gamma \mathbb{E}[\revr{Y_{m,k}}+Z_{m,k+1}|\pii_m] \right\}, \forall \gamma\geq 0. \label{eq:discount}
\end{align} 
 
From \cite{puterman2014markov}, we can establish the asymptotic equivalence between the average formulation and the discounted formulation:
 \begin{lemma}[Asymptotic Equivalence \cite{puterman2014markov}]\label{L2}
Given the optimal quotient value $\gamma^*$, 
Problems \eqref{eq:opt-rl} and \eqref{eq:average-p} are asymptotically equivalent to \revr{reformulation as follows:}
 \begin{align}
  \pii_m^*  = \arg\mathop{\rm{minimize}}_{\pii_{m}}~\lim_{\delta\rightarrow 1} \mathbb{E}[\Delta_m^\delta|\pii_m,\gamma^*],\label{eq:discount-p}
 \end{align}
for all $m\in\mathcal{M}$, where $\pii_m^*$ is the optimal solution to \eqref{eq:average-p}.
 \end{lemma}
Therefore, the discounted reformulation in \eqref{eq:discount-p} serves as a good approximation of \eqref{eq:average} when $\delta$ approaches $1$.
Such an approximation provides us with a convention of designing new DRL algorithms for fractional MDP problems based on existing well-established DRL algorithms.
We will stick to the discounted reformulation for the rest of this paper.


\subsection{Fractional MDP}

We study the following general fractional MDP framework and drop index $m$ for the rest of this section.

\begin{definition}[Fractional MDP]\label{D1}
A fractional MDP is defined as $(\mathcal{S},\mathcal{A},P,c_N,c_D,\delta)$, where $\mathcal{S}$ and $\mathcal{A}$ are the finite sets of states and actions, respectively; $P$ is the transition distribution; $c_N$ and $c_D$ are the cost functions ,
\footnote{\rev{Since we aim at minimizing the time-average AoI, we consider minimizing long-term expected cost  in this work. 
}}
and $\delta$ is a discount factor. We use $\mathcal{Z}$ to denote the joint state-action space, i.e., $\mathcal{Z}\triangleq\mathcal{S}\times \mathcal{A}$.
\end{definition}
From Definition \ref{D1} and Lemmas \ref{L1} and \ref{L2},
we have that Problem \eqref{eq:opt-rl} has the equivalent Dinkelbach's reformulation:
\begin{align}
{\bs\pi}^*=\arg\mathop{\text{minimize}}_{{\bs\pi}}\lim_{K\rightarrow \infty} \mathbb{E}\left[\left.\sum_{k=1}^K\delta^k ( c_N- \gamma^* c_D)\right|{\bs\pi}\right], \label{Din}
\end{align}
where we can see from Lemmas \ref{L1} and \ref{L2} that  $\gamma^*$ satisfies
\begin{align}
    \gamma^*=\mathop{\text{minimize}}_{{\bs\pi}}\lim_{K\rightarrow \infty}\frac{\left.\mathbb{E}\left[\sum_{k=0}^K\delta^k c_N\right|\pii\right]}{\mathbb{E}\left[\left.\sum_{k=0}^K\delta^k c_D\right|\pii\right]},\label{eq-FRL-2}
\end{align}

Note that Problem \eqref{Din} is a classical MDP problem, \revr{including} an immediate cost, given by $c_N(\bs{s},\bs{a})-\gamma^*c_D(\bs{s},\bs{a})$. Thus, we can then apply a traditional RL algorithm to solve such a reformulated problem, such as Q-Learning or its variants (e.g., SQL in \cite{SQL}).

However, the optimal quotient coefficient $\gamma^*$ and the transition distribution $P$ are unknown \textit{a priori}. Therefore, one needs to design an algorithm that combines both fractional programming  and RL algorithms to 
solve Problem \eqref{Din} for a given $\gamma$ and seek the value of $\gamma^*$. To achieve this, we start by introducing the following definitions:
    Given a quotient coefficient $\gamma$, the optimal Q-function is
    \begin{align}
        Q^*_\gamma(\bs{s},\bs{a})\triangleq \min_{\pii} Q^{\pii}_\gamma(\bs{s},\bs{a}),~\forall (\bs{s},\bs{a})\in\mathcal{Z}, \label{optimalQ}
    \end{align}
where $Q^{\pii}_\gamma(\bs{s},\bs{a})$ is the action-state function that satisfies the following Bellman's equation: for all $(\bs{s},\bs{a})\in\mathcal{Z}$,
\begin{align}
    Q^{\bs\pi}_\gamma(\bs{s},\bs{a})\triangleq c_N(\bs{s},\bs{a})-\gamma c_D(\bs{s},\bs{a})+\delta\mathbb{E}[Q^{\pii}_\gamma(\bs{s},\bs{a})|\bs\pi].
\end{align}
In addition,  we can further decompose the optimal Q-function in \eqref{optimalQ} into the following two parts: $ Q^*_\gamma(\bs{s},\bs{a})=N_\gamma(\bs{s},\bs{a})-\gamma D_\gamma(\bs{s},\bs{a})$
and, for all $(\bs{s},\bs{a})\in\mathcal{Z}$, 
\begin{align}
    N_\gamma(\bs{s},\bs{a})&=c_N(\bs{s},\bs{a})+ \delta\mathbb{E}[N_\gamma(\bs{s},\bs{a})|\bs\pi^*],\\ D_\gamma(\bs{s},\bs{a})&=c_D(\bs{s},\bs{a})+ \delta\mathbb{E}[ D_\gamma(\bs{s},\bs{a})|\bs\pi^*].
\end{align}





    
\begin{figure}[t]
	\begin{minipage}{\linewidth}
		\begin{algorithm}[H]
			\caption{Fractional Q-Learning (FQL)}\label{Algo-FQL}
			\small
			\begin{algorithmic}[1]
				\FOR{$k=1, 2,\ldots, K$} 
				\STATE Initialize $\s_{m}(1)$;\
				\FOR{time slot $t\in\T$}
				\STATE Observe the next state $\s_{m}(t+1)$;\
				\STATE Observe a set of costs $\{c_{m}(t'),~t'\in\widetilde{\T}_{m,t}\}$;\
				\FOR{each task $\task_m(t')$ with $t'\in\widetilde{\T}_{m,t}$}
				\STATE Send $(\s_{m}(t'),\a_{m}(t'),c_{m}(t'),\s_{m}(t'+1))$ to $n_{m}$;\
				\ENDFOR
				\ENDFOR
				\STATE $\gamma^{(k+1)}={N_{\gamma^{k}}(s,a_k)}/{D_{\gamma^k}(s,a_k)},$
				    where $a_k=\arg\max_{a} Q_{\gamma}^T(s,a)$.
				\ENDFOR
			\end{algorithmic}
		\end{algorithm}
	\end{minipage}
\end{figure}

\subsection{Fractional Q-Learning Algorithm}
In this subsection, we present a Fractional Q-Learning (FQL) algorithm in Algorithm \ref{Algo-FQL}, consisting of an inner loop with $E$ episodes and an outer loop. The key idea is to approximate the Q-function $Q_{\gamma_i}^*$ by $Q_i$ and then  iterate $\{\gamma_i\}$.

One of the key innovations in Algorithm \ref{Algo-FQL} is the design of the stopping condition, leading to the shrinking values of the uniform approximation errors of $Q_i$. This facilitates us to adapt the convergence proof in \cite{dinkelbach67ms} to our setting and prove
the linear convergence rate of $\{\gamma_i\}$
without increasing the inner-loop time complexity.

We describe the details of the inner loop and the outer loop procedures of Algorithm \ref{Algo-FQL} in the following:
\begin{itemize}
    \item \textit{Inner loop}: For each episode $i$, given a quotient coefficient $\gamma_i$, we perform an (arbitrary) Q-Learning algorithm (as the Speedy Q-Learning in \cite{SQL}) to approximate function $Q_{\gamma_i}^*(\bs{s},\bs{a})$ by $Q_i(\bs{s},\bs{a})$. Let $\s_0$ denote the initial state of any arbitrary episode, and $\bs{a}_i\triangleq \arg\max_{\bs{a}} Q_{i}(\bs{s}_0,\bs{a})$ for all $i\in[E]\triangleq\{1,...,E\}$. We consider a \textit{stopping condition} 
    \begin{align}
         \epsilon_i< -\alpha Q_i(\bs{s}_0,\bs{a}_i),~\forall i\in[E], \label{SC}
    \end{align}
 so as to terminate each episode $i$ with a bounded \textit{uniform approximation error}:
      $ \norm{Q^*_{\gamma_i}-Q_i}\leq \epsilon_i,~\forall i\in[E].$
    Operator $\norm{\cdot}$ is the supremum norm, which satisfies $ \lVert g \rVert\triangleq \max_{(\bs{s},\bs{a})\in\mathcal{Z}}g(\bs{s},\bs{a})$. 
    Specifically, we obtain $ Q_i(\bs{s},\bs{a})$, $N_i(\bs{s},\bs{a})$, and $D_i(\bs{s},\bs{a})$, which satisfy, for all $(\bs{s},\bs{a})\in\mathcal{Z}$,
    \begin{align}
      Q_i(\bs{s},\bs{a})=N_i(\bs{s},\bs{a})-\gamma_iD_i(\bs{s},\bs{a}).  \label{Eq-Qup}
    \end{align}

    \item \textit{ the outer loop to update \revr{the quotient coefficient:} }
    \begin{align}
        \gamma_{i+1}=\frac{N_i(\bs{s}_0,\bs{a}_i)}{D_i(\bs{s}_0,\bs{a}_i)}, \quad\forall i\in[E], \label{Update-O}
    \end{align}
   which will be shown to converge to the optimal value $\gamma^*$. 
\end{itemize}

\subsection{Convergence Analysis}

We are ready to present the key convergence results of our proposed FQL algorithm (Algorithm \ref{Algo-FQL}) as follows.

\subsubsection{Convergence of the outer loop}
We start with analyzing the convergence of the outer loop:
\begin{theorem}[Linear Convergence of Fractional Q-Learning]\label{T1}
If we select $\{T_i\}$ such that the uniform approximation error $\norm{Q^*_{\gamma_i}-Q_i}\leq \epsilon_i$ holds with $\epsilon_i< -\alpha Q_i(\bs{s}_0,\bs{a}_i)$ for  some $\alpha\in(0,1)$ and for all $i\in[E]$, then the sequence $\{\gamma_i\}$ generated by Algorithm \ref{Algo-FQL}  satisfies
\begin{equation}
    \frac{\gamma_{i+1}-\gamma^*}{\gamma_i-\gamma^*}\in(0,1),~\forall i\in[E] \text{ and }
    \lim_{i\rightarrow \infty} \frac{\gamma_{i+1}-\gamma^*}{\gamma_i-\gamma^*}=\alpha.
\end{equation}
That is, $\{\gamma_i\}$ converges to $\gamma^*$ linearly.
\end{theorem}
While the convergence proof in 
\cite{dinkelbach67ms} requires to obtain the \textit{exact} solution in each episode, Theorem \ref{T1} generalizes this result to the case where we only obtain an \textit{approximated (inexact)} solution in each episode. 
In addition to the proof techniques in \cite{dinkelbach67ms} and \cite{SQL},  our proof techniques include induction and
exploiting the convexity of $Q_i(\bs{s},\bs{a})$.
We present a proof sketch of Theorem \ref{T1} in \revr{Appendix A\footnote{Please refer to  \cite{agemec} for our appendices.}}.

The significance of Theorem \ref{T1} is two-fold. First, Theorem \ref{T1} shows that Algorithm \ref{Algo-FQL} achieves a linear convergence rate, even though it only attains an approximation of \revr{$Q_\gamma^*(\bs{s},\bs{a})$ \deleting{in each episode $i\in[E]$}}. Second,  \eqref{SC} is a well-behaved stopping condition.

\subsubsection{Time Complexity of the inner loop}
Although as $\{Q_i(\bs{s}_0,\bs{a}_i)\}$ is convergent to $0$ and hence $\epsilon_i<-\alpha Q_i$ is getting more restrictive as $i$ increases, the steps needed $T_i$ in Algorithm \ref{Algo-FQL} keep to be finite without increasing over episode $i$. See Appendix B in detail.



\begin{figure}[t]
	\centering
 	\includegraphics[width=7cm]{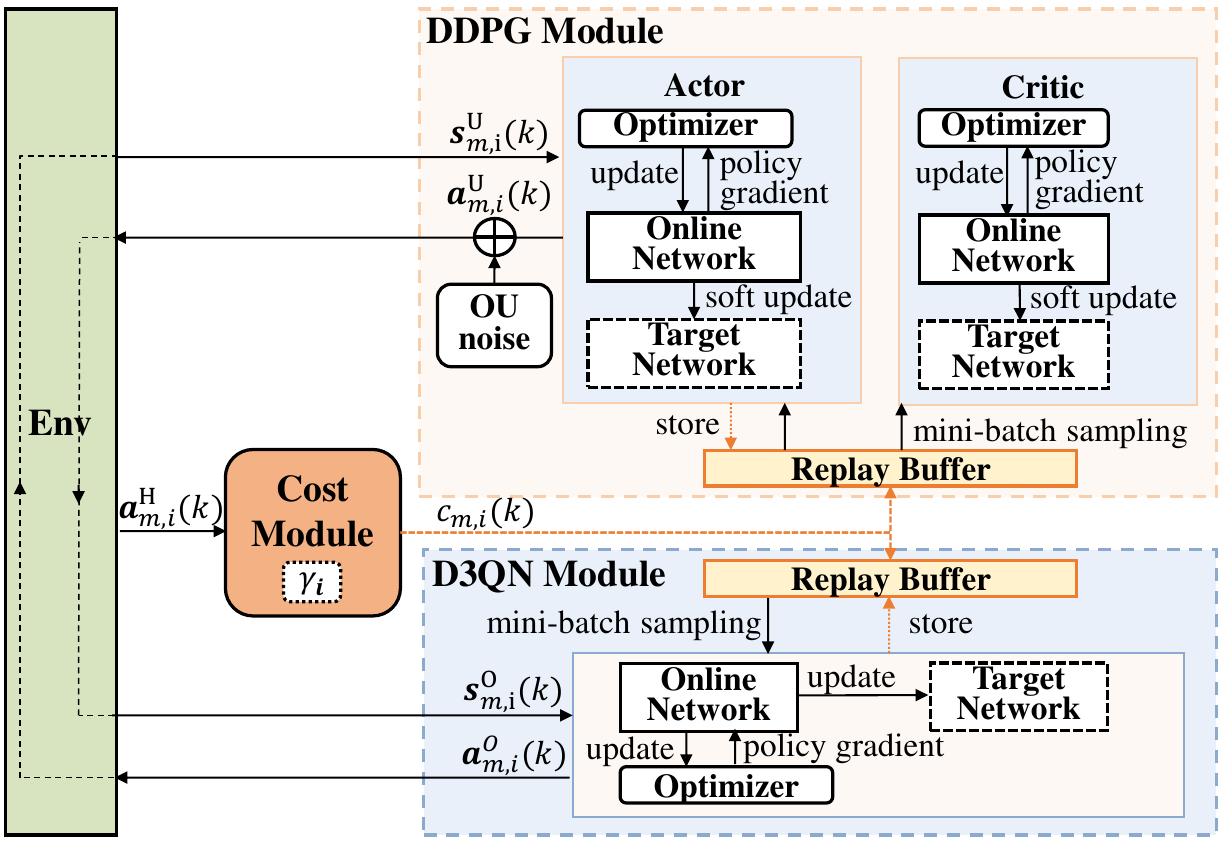}
	\caption{\revr{Illustration on the proposed fractional DRL framework.}
	}
 \label{fig:DNN}
\end{figure}

\section{Fractional DRL Algorithm}\label{sec:algorithm}

{In this section, we present a fractional DRL-based algorithm to \rev{approximate the Q-function in FQL algorithm and} solve Problem \eqref{eq:opt-rl} with DDPG (for continuous action space) \cite{lillicrap2015continuous} and D3QN (for discrete action space) \cite{mnih2015human} techniques
\revr {for the task updating and offloading processes in the decentralized manner, which is illustrated in Fig. \ref{fig:DNN}. Appendix C shows detailed settings of networks. }Moreover, we design a cost function based on \revr {our} fractional RL framework to ensure the convergence. 

\subsection{Cost Module}
As in the proposed fractional RL framework, we consider a set of episodes $i\in[E]$ and introduce a quotient coefficient $\gamma_i$ for episode $i$. Consider mobile device $m$. Let 
$  \ah_{m,i}(k)\triangleq \{(\au_{m,i}(l), \ao_{m,i}(l))\}_{l=1}^k$
denote the set of  updating and offloading actions of mobile device $m\in\M$ made until task $k$  in episode $i$, where ``H" refers to ``history". Recall that $\s_0$ is the initial state of any arbitrary episode. 
We define $N_i(\s_0,\ah_{m,i})$ and $D_i(\s_0, \ah_{m,i})$ as follows:
\begin{align}
N_i(\s_0,\ah_{m,i}(k)) &=  \sum_{l=1}^k A(Y_{m,l}^i,Z_{m,l+1}^i,Y_{m,l+1}^i),\label{eq:N}\\
   D_i(\s_0, \ah_{m,i}(k)) &= \sum_{l=1}^k ({Y_{m,l}^i}+Z_{m,l+1}^i),\label{eq:D}
\end{align}
{where $Y_{m,l}^i$ and $Z_{m,l+1}^i$ are the delay of task $l$ and \revr{the wait interval} for generating the next task $l+1$, respectively, for mobile device $m$. Note that $Y_{m,l}^i$ is a function of $\ah_{m,i}(l)$, and $Z_{m,l+1}^i$ is a function of $\ah_{m,i}(l)$ and $\au_{m,i}(l+1)$. The cost module keeps track of $\ah_{m,i}(k)$ or equivalently $Y^i_{m,l}$ and $Z^i_{m,l}$ for all $l=1,...,k$ across the training process.} 

In step $k$ of episode $i$,
a cost is determined and sent to the DDPG and D3QN modules. This process corresponds to the inner loop of the proposed fractional RL framework and is defined based on \eqref{eq:N}: for all $i\in[E], m\in\mathcal{M}, k\in\mathcal{K},$
\begin{align}
    c_{m,i}(k)\!=\!A(Y_{m,k}^i,Z_{m,k+1}^i,&Y_{m,k+1}^i)\!-\! \gamma_{m,i} \! \cdot\! (Y_{m,k}^i\!+\!Z_{m,k+1}^i), 
    \label{eq:cost-DRL}
\end{align}
where $A(Y_{m,k}^i,Z_{m,k+1}^i,Y_{m,k+1}^i)$ stands for the area of a trapezoid in \eqref{trapezoid}. The cost in \eqref{eq:cost-DRL} corresponds to an (immediate) cost function as in the fractional MDP problem in \eqref{Din}. 


Finally, at the end of each episode $i$, the cost module updates $\gamma_{m,i+1}$ using \eqref{eq:N} and \eqref{eq:D}:
 \begin{align}
    \gamma_{m,i+1}=\frac{N_i(\s_0,\ah_{m,i}(T))}{D_i(\s_0,\ah_{m,i}(T))},~~ i\in[E], \label{eq:gammaupdate}
\end{align}
where $T$ is the stepped needed set to be the same for every episode, as motivated in Appendix B. Eq. \eqref{eq:gammaupdate}
corresponds to the update procedure of the quotient coefficient in \eqref{Update-O} as in the outer loop of the fractional RL framework.

 
\deleting{In the following, we will describe the details of the DDPG and  D3QN modules. For presentation simplicity, we focus on one episode and hence drop the episode index $i$. }

\begin{figure}[t]
\centering
\includegraphics[height=3.3cm]{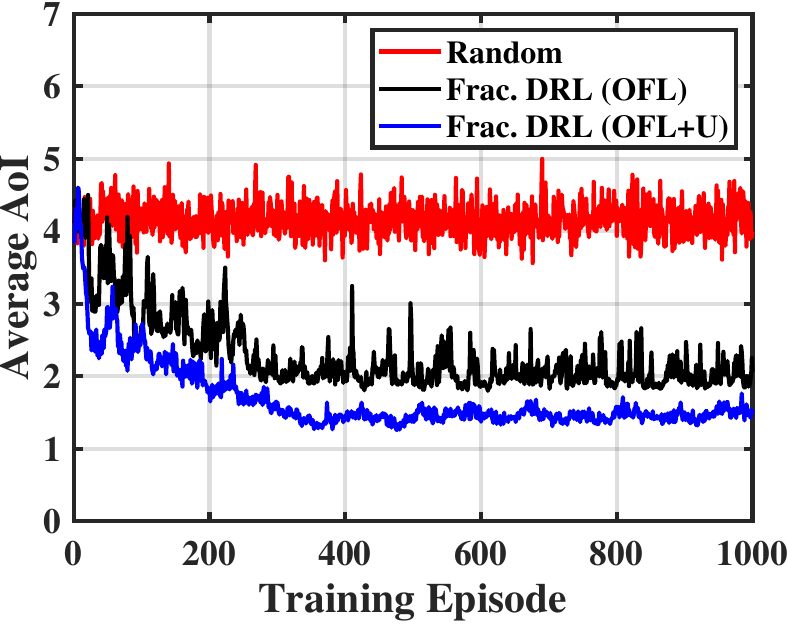}\label{fig:aoi}
\includegraphics[height=3.3cm]{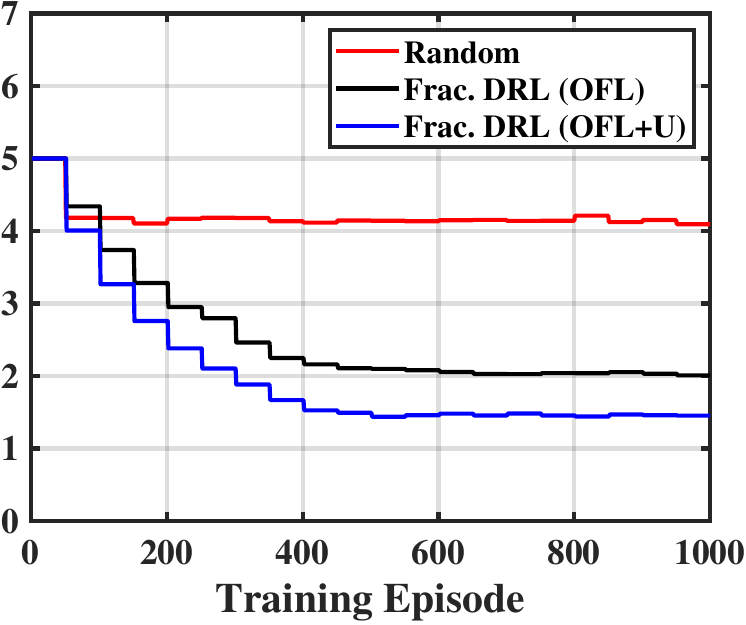}\label{fig:gamma}\\
\quad(a)\qquad\qquad\qquad\qquad\qquad\qquad(b)
\caption{
Convergence of (a)  average AoI and (b) the average value of quotient coefficients $\gamma_{m,i}$ across devices, where $\gamma_{m,i}$ is updated every $50$ episodes.}
\label{fig:convergence}
\end{figure}

\begin{figure*}[t]
\centering
\includegraphics[height=3.25cm]{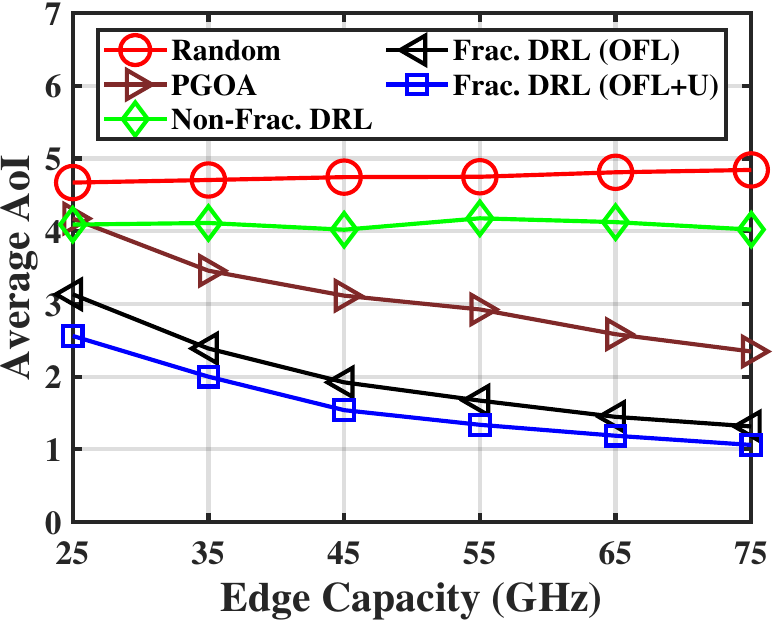}\label{fig:ratio}
\includegraphics[height=3.25cm]{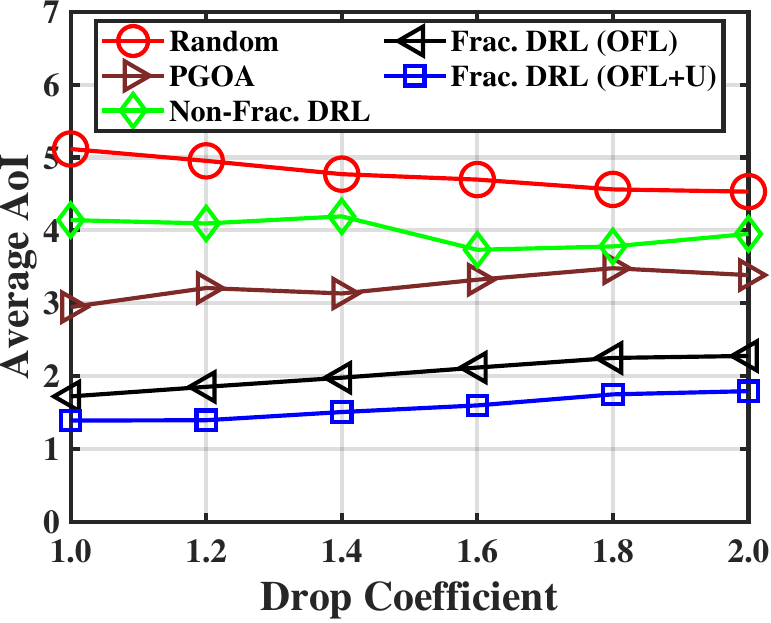}\label{fig:drop}
\includegraphics[height=3.2cm]{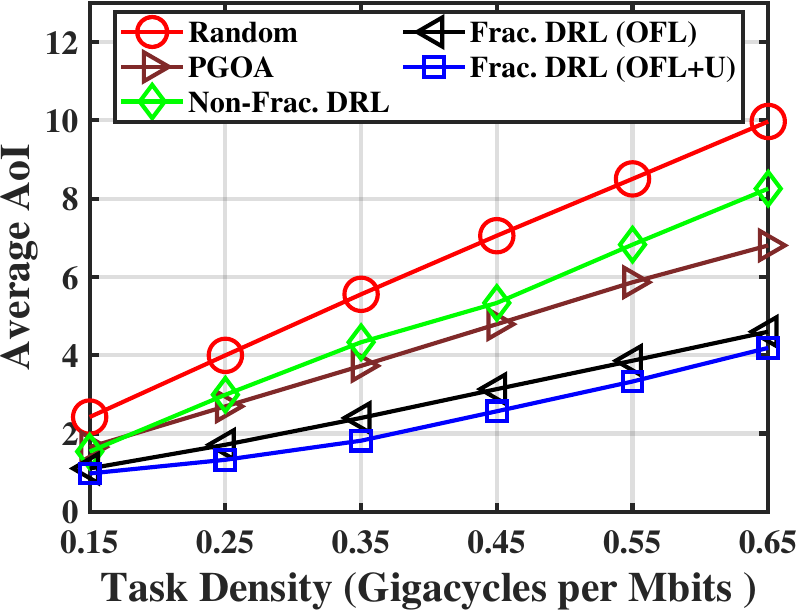}\label{fig:density}
\includegraphics[height=3.2cm]{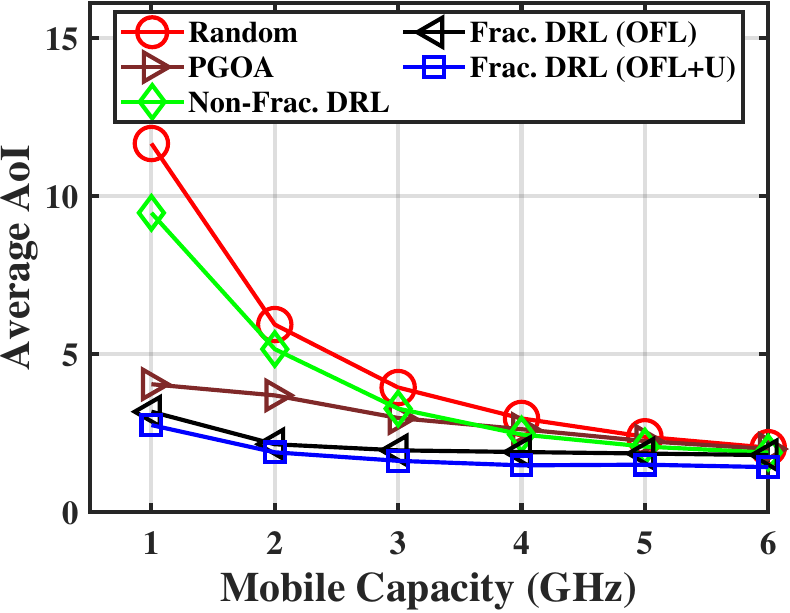}\label{fig:capacity}\\
(a)\qquad\qquad\qquad\qquad\qquad\qquad(b)\qquad\qquad\qquad\qquad\qquad(c)\qquad\qquad\qquad\qquad\qquad(d)

\caption{
Performance  under different  (a) processing capacities of edges,
(b) drop coefficients, (c) task densities, and (d) processing capacities of mobile devices.}
\label{fig:compare}
\end{figure*}

\section{Performance Evaluation}\label{sec:experiment}

We perform experiments to evaluate our proposed fractional DRL algorithm. \rev{We consider two edge nodes and 20 mobile devices \rev{ learning their own policies simultaneously}. Unless otherwise specified, we follow the experimental settings in \cite[Table I]{Tang2020TMC}.} 
\revr {We present more detailed experiment settings in \revr{Appendix D}.}


\rev{We denote proposed Frac. DRL (OFL+U), which is short for ``fractional DRL with offloading and updating policies".} This is compared with several benchmark methods: 
\begin{itemize}
    \item \textit{Random scheduling}: The updating and offloading decisions are randomly generated within action space.
    \item \textit{PGOA \cite{yang2018distributed}}: \rev{This corresponds to a best response algorithm for potential game in MEC systems.} 
    \item \textit{Non-Fractional DRL} (denoted by Non-Frac. DRL) \cite{xu2022aoi}: \rev{This benchmark adopts D3QN network to learn the offloading policy. In contrast to our proposed framework, this benchmark is non-fractional. That is, its objective} approximates the ratio-of-expectation average AoI in \eqref{Problem-fractional} by an \textit{expectation-of-ratio} expression:
$\mathop{\text{minimize}}_{\pii_{m}}~~ \mathbb{E}\left[\left.\frac{ A(Y_{m,k},Z_{m,k+1},Y_{m,k+1})}{ Y_{m,k}+Z_{m,k+1}}\right|\pii_m\right].$
Such an approximation can circumvent the fractional challenge but incurs large accuracy loss. 
 \item \rev{\textit{Frac. DRL with Offloading Only} (denoted by Frac. DRL (OFL)): We propose this algorithm by simplifying our fractional DRL algorithm through considering only the offloading policy. The updating policy (i.e., $Z_{m,k}$) is set to zero, as in many of the existing works \cite{xie2022reinforcement,he2022age,xu2022aoi}.} 
\end{itemize}
\rev{The performance difference between Non-Frac. DRL and Frac. DRL (OFL) shows the significance of our proposed fractional DRL. The difference between Frac. DRL (OFL) and Frac. DRL (OFL+U) shows the necessity of joint offloading and waiting optimization.}

\textbf{Convergence}: Fig. \ref{fig:convergence} illustrates the convergence of  our proposed Frac. DRL (OFL) and Frac. DRL (OFL+U) algorithms. \rev{Unlike non-fractional approaches, our proposed approach involves the convergence of not only the neural network (see Fig. \ref{fig:convergence}(a)) but also the quotient coefficient $\gamma$ (see Fig. \ref{fig:convergence}(b)). As a result, the convergence curve of AoI may sometimes change non-monotonically. In  Fig. \ref{fig:convergence}(a), both Frac. DRL (OFL) and Frac. DRL (OFL+U) converge after roughly 350 episodes. As for the converged AoI, \revr{Frac. DRL (OFL+U)  outperforms Frac. DRL (OFL) by 31.3\%.}} 






\textbf{Edge Capacity}: Fig. \ref{fig:compare}(a) evaluates the performance of our proposed schemes under different node processing capacities. 
First, the proposed fractional DRL-based algorithm consistently achieves lower average AoI, compared against those non-fractional benchmarks. Such an advantage is more significant as the per-node processing capacity is larger. When processing capacity is $75$ GHz, Frac. DRL (OFL+U)  can achieve an average AoI \deleting{reduction of $72.8\%$ and $72.2\%$}\revr{reduction of $54.8\%$ and $73.6\%$, compared against PGOA and Non-Frac. DRL, respectively.} 
Second, when compared with Frac. DRL (OFL), Frac. DRL (OFL+U) can further reduce the average AoI  \revr{up to $19.9\%$} when the processing capacity of edge nodes is 55 GHz. This further shows that
a well-designed updating policy also plays an important role of further improving the performance, especially there are relatively high edge loads.

\textbf{Drop Coefficient}: In Fig. \ref{fig:compare}(b), we consider different drop coefficients, i.e., the ratio of the drop time $\bar{Y}$ to the average time of processing a task. 
The performance gaps between the proposed schemes and benchmarks are large when the drop ratio gets smal (i.e., tasks are more delay-sensitive). When the drop coefficient is \revr{1.0}, the Frac. DRL (OFL+U) reduces the average AoI \revr{by $57.6\%$}, compared with Non-Frac. DRL. 

\textbf{Task Density}: In Fig. \ref{fig:compare}(c), we evaluate algorithm performance under different task densities, which affect the expected processing time of tasks at both edge nodes and mobile devices. Specifically, our proposed  Frac. DRL (OFL) and Frac. DRL (OFL+U) schemes outperform all the benchmarks. In addition, the performance gaps increase as the task density increases, which shows the benefit of \revr{our proposed algorithm} under large task densities. When the task density is 0.65, Frac. DRL (OFL+U) achieves an average \deleting{reductions of  $55.0\%$ and $54.5\%$}\revr{reductions of  $38.4\%$ and $49.3\%$}, compared against PGOA and Non-Frac. DRL, respectively. Meanwhile, Frac. DRL (OFL+U) outperforms Frac. DRL (OFL) \revr{by up to $24.2\%$}. 

\textbf{Mobile Capacity}: In Fig. \ref{fig:compare}(d), as the processing capacity of mobile devices decreases, the gap between Frac. DRL (OFL) and Non-Frac. DRL significantly increases, indicating the necessity of our fractional scheme. When mobile capacity is 2 GHz, Frac. DRL (OFL+U) can achieve an average AoI reduction of 48.8\%  compared to PGOA.  



To summarize, \textit{our proposed schemes significantly outperform non-fractional benchmarks, especially under large task density, delay-sensitive tasks, and small mobile device processing capacity.  Meanwhile, the joint optimization over offloading and updating can further increase the system performance \deleting{by up to $31.4$\%.}\revr{by up to $31.3$\%.}} We present additional convergence and performance evaluation under different networks hyperparamters, distribution of processing duration and scale of mobile devices in Appendix D.


\section{Conclusion}\label{sec:conclusion}

This paper has studied the computational task scheduling  (including offloading and updating) problem for age-minimal MEC. To address the underlying challenges of unknown load dynamics and  the fractional objective, we have proposed a fractional RL framework with a provable linear convergence rate.
We  further designed a fractional DRL algorithm that incorporates D3QN and DDPG techniques to tackle hybrid action space.
Experimental results show that proposed fractional algorithms significantly reduce the average AoI, compared against several benchmarks. Meanwhile, the joint optimization of offloading and updating can further reduce the average AoI,  validating the effectiveness of our proposed  scheme. There are several future directions, including incorporating multi-agent RL with recurrent neural networks for non-stationarity and social optimal scheduling. 

\section{Acknowledgments}
This work was supported in part by the National Natural Science Foundation of China (Projects 62202427 and 62202214), in part by Guangdong Basic and Applied Basic Research Foundation under Grant 2023A1515012819, and in part by the Australian Research Council (ARC) Discovery Early Career Researcher Award (DECRA) under Grant DE230100046.

\bibliography{aaai24}

\begin{thebibliography}{48}
\providecommand{\natexlab}[1]{#1}

\bibitem[{Akbari et~al.(2021)}]{AoIDeep2}
Akbari, M.; et~al. 2021.
\newblock Age of Information Aware VNF Scheduling in Industrial IoT Using Deep Reinforcement Learning.
\newblock \emph{IEEE J. Sel. Areas Commun.}, 39(8): 2487--2500.

\bibitem[{Ceran, G{\"u}nd{\"u}z, and Gy{\"o}rgy(2021)}]{ceran2021reinforcement}
Ceran, E.~T.; G{\"u}nd{\"u}z, D.; and Gy{\"o}rgy, A. 2021.
\newblock A reinforcement learning approach to age of information in multi-user networks with {HARQ}.
\newblock \emph{IEEE J. Sel. Areas Commun.}, 39(5): 1412--1426.

\bibitem[{Chen and Xie(2022)}]{Chen_Xie_2022}
Chen, J.; and Xie, H. 2022.
\newblock An Online Learning Approach to Sequential User-Centric Selection Problems.
\newblock \emph{Proceedings of the AAAI Conference on Artificial Intelligence}, 36(66): 6231–6238.

\bibitem[{Chen et~al.(2020)}]{AoIDeep1}
Chen, X.; et~al. 2020.
\newblock Age of Information Aware Radio Resource Management in Vehicular Networks: A Proactive Deep Reinforcement Learning Perspective.
\newblock \emph{IEEE Trans. Wireless Commun.}, 19(4).

\bibitem[{Chen et~al.(2022)}]{chen2022info}
Chen, X.; et~al. 2022.
\newblock Information Freshness-Aware Task Offloading in Air-Ground Integrated Edge Computing Systems.
\newblock \emph{IEEE J. Sel. Areas Commun.}, 40(1): 243--258.

\bibitem[{Chiariotti et~al.(2021)Chiariotti, Vikhrova, Soret, and Popovski}]{AoIMEC1}
Chiariotti, F.; Vikhrova, O.; Soret, B.; and Popovski, P. 2021.
\newblock Peak Age of Information Distribution for Edge Computing With Wireless Links.
\newblock \emph{IEEE Trans. Commun.}, 69(5): 3176--3191.

\bibitem[{Dinkelbach(1967)}]{dinkelbach67ms}
Dinkelbach, W. 1967.
\newblock On nonlinear fractional programming.
\newblock \emph{Management science}, 13(7): 492--498.

\bibitem[{Ghavamzadeh et~al.(2011)Ghavamzadeh, Kappen, Azar, and Munos}]{SQL}
Ghavamzadeh, M.; Kappen, H.; Azar, M.; and Munos, R. 2011.
\newblock Speedy Q-Learning.
\newblock In \emph{Proc. Neural Info. Process. Syst. (NIPS)}, volume~24.

\bibitem[{He et~al.(2022)He, Wang, Wang, Xu, and Ren}]{he2022age}
He, X.; Wang, S.; Wang, X.; Xu, S.; and Ren, J. 2022.
\newblock Age-Based Scheduling for Monitoring and Control Applications in Mobile Edge Computing Systems.
\newblock In \emph{Proc. IEEE INFOCOM}.

\bibitem[{Hu et~al.(2020)Hu, Zhang, Song, Schober, and Poor}]{UAVDeep1}
Hu, J.; Zhang, H.; Song, L.; Schober, R.; and Poor, H.~V. 2020.
\newblock Cooperative {I}nternet of {UAV}s: Distributed Trajectory Design by Multi-Agent Deep Reinforcement Learning.
\newblock \emph{IEEE Trans. Commun.}

\bibitem[{Huang, Bi, and Zhang(2020)}]{huang2020deep}
Huang, L.; Bi, S.; and Zhang, Y.-J.~A. 2020.
\newblock Deep Reinforcement Learning for Online Computation Offloading in Wireless Powered Mobile-Edge Computing Networks.
\newblock \emph{IEEE Trans. Mobile Comput.}, 19(11): 2581 -- 2593.

\bibitem[{Jin et~al.(2023)Jin, Tang, Zhang, and Wang}]{agemec}
Jin, L.; Tang, M.; Zhang, M.; and Wang, H. 2023.
\newblock Fractional Deep Reinforcement Learning for Age-Minimal Mobile Edge Computing.

\bibitem[{Kaul, Yates, and Gruteser(2012)}]{kaul2012real}
Kaul, S.; Yates, R.; and Gruteser, M. 2012.
\newblock Real-time status: How often should one update?
\newblock In \emph{Proc. IEEE INFOCOM}, 2731--2735.

\bibitem[{Keating and Shadwick(2002)}]{keating2002universal}
Keating, C.; and Shadwick, W.~F. 2002.
\newblock A universal performance measure.
\newblock \emph{Journal of performance measurement}, 6(3): 59--84.

\bibitem[{Kuang et~al.(2020)Kuang, Gong, Chen, and Ma}]{AoIMEC4}
Kuang, Q.; Gong, J.; Chen, X.; and Ma, X. 2020.
\newblock Analysis on computation-intensive status update in mobile edge computing.
\newblock \emph{IEEE Transactions on Vehicular Technology}, 69(4): 4353--4366.

\bibitem[{Li, Zhou, and Chen(2020)}]{li2020age}
Li, J.; Zhou, Y.; and Chen, H. 2020.
\newblock Age of Information for Multicast Transmission With Fixed and Random Deadlines in {IoT} Systems.
\newblock \emph{IEEE Internet of Things Journal}, 7(9): 8178--8191.

\bibitem[{Lillicrap et~al.(2015)Lillicrap, Hunt, Pritzel, Heess, Erez, Tassa, Silver, and Wierstra}]{lillicrap2015continuous}
Lillicrap, T.~P.; Hunt, J.~J.; Pritzel, A.; Heess, N.; Erez, T.; Tassa, Y.; Silver, D.; and Wierstra, D. 2015.
\newblock Continuous control with deep reinforcement learning.
\newblock \emph{arXiv preprint arXiv:1509.02971}.

\bibitem[{Liu et~al.(2022{\natexlab{a}})Liu, Zheng, Huang, and Quek}]{liu2022distributed}
Liu, S.; Zheng, C.; Huang, Y.; and Quek, T. Q.~S. 2022{\natexlab{a}}.
\newblock Distributed Reinforcement Learning for Privacy-Preserving Dynamic Edge Caching.
\newblock \emph{IEEE J. Sel. Areas Commun.}, 40(3): 749--760.

\bibitem[{Liu et~al.(2022{\natexlab{b}})}]{liu2022deep}
Liu, T.; et~al. 2022{\natexlab{b}}.
\newblock Deep Reinforcement Learning based Approach for Online Service Placement and Computation Resource Allocation in Edge Computing.
\newblock \emph{IEEE Trans. Mobile Comput.}

\bibitem[{M.~Zhou and Yates(2024, Early Access)}]{zhou2024}
M.~Zhou, H. H.~Y., M.~Zhang; and Yates, R.~D. 2024, Early Access.
\newblock Age-minimal CPU Scheduling.
\newblock In \emph{Proc. IEEE INFOCOM}.

\bibitem[{Ma et~al.(2022)Ma, Huang, Zhou, Zhang, and Chen}]{ma2022green}
Ma, H.; Huang, P.; Zhou, Z.; Zhang, X.; and Chen, X. 2022.
\newblock GreenEdge: Joint Green Energy Scheduling and Dynamic Task Offloading in Multi-Tier Edge Computing Systems.
\newblock \emph{IEEE Transactions on Vehicular Technology}, 71(4): 4322--4335.

\bibitem[{Mao et~al.(2017)Mao, You, Zhang, Huang, and Letaief}]{mao2017survey}
Mao, Y.; You, C.; Zhang, J.; Huang, K.; and Letaief, K.~B. 2017.
\newblock A survey on mobile edge computing: The communication perspective.
\newblock \emph{IEEE Commun. Surveys \& Tuts.}, 19(4): 2322--2358.

\bibitem[{Mnih et~al.(2015)}]{mnih2015human}
Mnih, V.; et~al. 2015.
\newblock Human-level control through deep reinforcement learning.
\newblock \emph{Nature}, 518(7540): 529--533.

\bibitem[{Omidkar et~al.(2022)Omidkar, Khalili, Nguyen, and Shafiei}]{9709888}
Omidkar, A.; Khalili, A.; Nguyen, H.~H.; and Shafiei, H. 2022.
\newblock Reinforcement-Learning-Based Resource Allocation for Energy-Harvesting-Aided D2D Communications in IoT Networks.
\newblock \emph{IEEE Internet Things J}, 9(17): 16521--16531.

\bibitem[{Porambage et~al.(2018)}]{porambage2018survey}
Porambage, P.; et~al. 2018.
\newblock Survey on multi-access edge computing for {Internet of things} realization.
\newblock \emph{IEEE Commun. Surveys \& Tuts.}, 20(4): 2961--2991.

\bibitem[{Puterman(2014)}]{puterman2014markov}
Puterman, M.~L. 2014.
\newblock \emph{Markov decision processes: discrete stochastic dynamic programming}.
\newblock John Wiley \& Sons.

\bibitem[{Ren and Krogh(2005)}]{1431043}
Ren, Z.; and Krogh, B. 2005.
\newblock Markov decision Processes with fractional costs.
\newblock \emph{IEEE Transactions on Automatic Control}, 50(5): 646--650.

\bibitem[{Shisher and Sun(2022)}]{shesher2022how}
Shisher, M. K.~C.; and Sun, Y. 2022.
\newblock How Does Data Freshness Affect Real-Time Supervised Learning?
\newblock In \emph{Proc. ACM MobiHoc}. Seoul, Republic of Korea.

\bibitem[{Sun et~al.(2017)Sun, Uysal-Biyikoglu, Yates, Koksal, and Shroff}]{sun2017tit}
Sun, Y.; Uysal-Biyikoglu, E.; Yates, R.~D.; Koksal, C.~E.; and Shroff, N.~B. 2017.
\newblock Update or Wait: How to Keep Your Data Fresh.
\newblock \emph{IEEE Trans. Inform. Theory}, 63(11): 7492--7508.

\bibitem[{Suttle et~al.(2021)Suttle, Zhang, Yang, Liu, and Kraemer}]{pmlr-v139-suttle21a}
Suttle, W.; Zhang, K.; Yang, Z.; Liu, J.; and Kraemer, D. 2021.
\newblock Reinforcement Learning for Cost-Aware Markov Decision Processes.
\newblock In Meila, M.; and Zhang, T., eds., \emph{Proceedings of the 38th International Conference on Machine Learning}, volume 139 of \emph{Proceedings of Machine Learning Research}, 9989--9999. PMLR.

\bibitem[{Taka, He, and Oki(2022)}]{taka2022service}
Taka, H.; He, F.; and Oki, E. 2022.
\newblock Service Placement and User Assignment in Multi-Access Edge Computing with Base-Station Failure.
\newblock In \emph{Proc. IEEE/ACM IWQoS}.

\bibitem[{Tanaka(2017)}]{doi:10.1080/02522667.2015.1105525}
Tanaka, T. 2017.
\newblock A partially observable discrete time Markov decision process with a fractional discounted reward.
\newblock \emph{Journal of Information and Optimization Sciences}, 38(1): 21--37.

\bibitem[{Tang and Wong(2022)}]{Tang2020TMC}
Tang, M.; and Wong, V.~W. 2022.
\newblock Deep Reinforcement Learning for Task Offloading in Mobile Edge Computing Systems.
\newblock \emph{IEEE Trans. Mobile Comput.}, 21(6): 1985--1997.

\bibitem[{Tang et~al.(2021)Tang, Sun, Yang, and Zhou}]{tang2021age}
Tang, Z.; Sun, Z.; Yang, N.; and Zhou, X. 2021.
\newblock Age of Information Analysis of Multi-user Mobile Edge Computing Systems.
\newblock In \emph{Proc. IEEE GLOBECOM}.

\bibitem[{Tuli et~al.(2022)}]{tuli2022dynamic}
Tuli, S.; et~al. 2022.
\newblock Dynamic Scheduling for Stochastic Edge-Cloud Computing Environments Using {A3C} Learning and Residual Recurrent Neural Networks.
\newblock \emph{IEEE Trans. Mobile Comput.}, 21(3).

\bibitem[{Wang et~al.(2021)Wang, Guo, Zhang, Yang, Zhou, and Shen}]{wang2021delay}
Wang, S.; Guo, Y.; Zhang, N.; Yang, P.; Zhou, A.; and Shen, X. 2021.
\newblock Delay-Aware Microservice Coordination in Mobile Edge Computing: A Reinforcement Learning Approach.
\newblock \emph{IEEE Trans. Mobile Comput.}, 20(3): 939 -- 951.

\bibitem[{Wang et~al.(2022{\natexlab{a}})}]{wang2022distributed}
Wang, S.; et~al. 2022{\natexlab{a}}.
\newblock Distributed Reinforcement Learning for Age of Information Minimization in Real-Time IoT Systems.
\newblock \emph{IEEE J. Sel. Top. Signal Process.}, 16(3): 501--515.

\bibitem[{Wang, Ye, and Lui(2022)}]{wang2022decentralized}
Wang, X.; Ye, J.; and Lui, J.~C. 2022.
\newblock Decentralized Task Offloading in Edge Computing: A Multi-User Multi-Armed Bandit Approach.
\newblock In \emph{Proc. IEEE Conference on Computer Communications (INFOCOM)}.

\bibitem[{Wang et~al.(2022{\natexlab{b}})Wang, Wei, Richard~Yu, and Han}]{wang2022utility}
Wang, Z.; Wei, Y.; Richard~Yu, F.; and Han, Z. 2022{\natexlab{b}}.
\newblock Utility Optimization for Resource Allocation in Multi-Access Edge Network Slicing: A Twin-Actor Deep Deterministic Policy Gradient Approach.
\newblock \emph{IEEE Trans. Wireless Commun.}, 1--1.

\bibitem[{Wu et~al.(2021)Wu, Zhang, Wu, Han, Poor, and Song}]{UAVDeep2}
Wu, F.; Zhang, H.; Wu, J.; Han, Z.; Poor, H.~V.; and Song, L. 2021.
\newblock {UAV}-to-Device Underlay Communications: Age of Information Minimization by Multi-Agent Deep Reinforcement Learning.
\newblock \emph{IEEE Transactions on Communications}, 69(7): 4461--4475.

\bibitem[{Xie, Wang, and Weng(2022)}]{xie2022reinforcement}
Xie, X.; Wang, H.; and Weng, M. 2022.
\newblock A Reinforcement Learning Approach for Optimizing the Age-of-Computing-Enabled IoT.
\newblock \emph{IEEE Internet of Things Journal}, 9(4): 2778--2786.

\bibitem[{Xu et~al.(2022)}]{xu2022aoi}
Xu, C.; et~al. 2022.
\newblock {AoI}-centric Task Scheduling for Autonomous Driving Systems.
\newblock In \emph{Proc. IEEE INFOCOM}.

\bibitem[{Yang et~al.(2018)Yang, Zhang, Li, Ji, and Leung}]{yang2018distributed}
Yang, L.; Zhang, H.; Li, X.; Ji, H.; and Leung, V. 2018.
\newblock A Distributed Computation Offloading Strategy in Small-Cell Networks Integrated With Mobile Edge Computing.
\newblock \emph{IEEE/ACM Trans. Netw.}, 26(6).

\bibitem[{Yates et~al.(2021)Yates, Sun, Brown, Kaul, Modiano, and Ulukus}]{AoISurvey2021}
Yates, R.~D.; Sun, Y.; Brown, D.~R.; Kaul, S.~K.; Modiano, E.; and Ulukus, S. 2021.
\newblock Age of Information: An Introduction and Survey.
\newblock \emph{IEEE J. Sel. Areas Commun.}, 39(5): 1183--1210.

\bibitem[{Zhao et~al.(2022)Zhao, Ye, Pei, Liang, and Niyato}]{zhao2022multi}
Zhao, N.; Ye, Z.; Pei, Y.; Liang, Y.-C.; and Niyato, D. 2022.
\newblock Multi-Agent Deep Reinforcement Learning for Task Offloading in {UAV}-assisted Mobile Edge Computing.
\newblock \emph{IEEE Trans. Wireless Commun.}

\bibitem[{Zhu and Gong(2022)}]{zhu2022online}
Zhu, J.; and Gong, J. 2022.
\newblock Online Scheduling of Transmission and Processing for AoI Minimization with Edge Computing.
\newblock In \emph{Proc. IEEE INFOCOM WKSHPS}.

\bibitem[{Zhu et~al.(2022)Zhu, Wan, Fan, and Letaief}]{zhu2022federated}
Zhu, Z.; Wan, S.; Fan, P.; and Letaief, K.~B. 2022.
\newblock Federated Multiagent Actor–Critic Learning for Age Sensitive Mobile-Edge Computing.
\newblock \emph{IEEE Internet of Things Journal}, 9(2): 1053--1067.

\bibitem[{Zou, Ozel, and Subramaniam(2021)}]{AoIMEC3}
Zou, P.; Ozel, O.; and Subramaniam, S. 2021.
\newblock Optimizing Information Freshness Through Computation–Transmission Tradeoff and Queue Management in Edge Computing.
\newblock \emph{IEEE/ACM Trans. Netw.}, 29(2).

\end{thebibliography}

\newpage

\section*{Appendix A: Proof of Theorem 1}\label{Proof-T1}

Define 
\begin{subequations}
\begin{align}
   \bs{a}^*(\gamma) &\triangleq \arg\max_{a\in\mathcal{A}}\left(N_{\gamma}(\bs{s}_0,\bs{a})-\gamma D_{\gamma}(\bs{s}_0,\bs{a})\right), \\
    Q({\gamma})& \triangleq \max_{\bs{a}\in\mathcal{A}}\left(N_{\gamma}(\bs{s}_0,\bs{a})-\gamma D_{\gamma}(\bs{s}_0,\bs{a})\right),\\
    N(\gamma)&=N_\gamma(s_0,\bs{a}^*(\gamma))~~{\rm and}~~D(\gamma)=D_\gamma(s_0,\bs{a}^*(\gamma)),\\
    F(\gamma)&=\frac{N(\gamma)}{D(\gamma)},\\
       \bs{a}_i &\triangleq \arg\max_{\bs{a}\in\mathcal{A}}\left(N_{k}(\bs{s}_0,\bs{a})-\gamma D_{k}(\bs{s}_0,\bs{a})\right),
\end{align}
\end{subequations}
for all $\gamma\geq 0$.
In the remaining part of this proof, we use $Q_i=Q_i(\bs{s}_0,\bs{a}_i)$, $N_i=N_i(\bs{s}_0,\bs{a}_i)$, and $D_i=D_i(\bs{s}_0,\bs{a}_i)$ for presentation simplicity.
Note that
\begin{align}
    &\frac{N(\gamma')}{D(\gamma')}-\frac{N_i}{D_i}\label{Proof-Eq1}\\
    \overset{(a)}{\geq}~&\frac{N(\gamma')}{D(\gamma')}-\frac{N(\gamma')}{D_i}-\gamma\left[\frac{D(\gamma')}{D_i}-\frac{D(\gamma')}{D(\gamma)}\right]-\frac{\epsilon_i}{D_i}\nonumber\\
    =~&[-Q(\gamma')+(\gamma_i-\gamma')D({\gamma'})]
    \left(\frac{1}{D_i}-\frac{1}{D(\gamma')}\right)-\frac{\epsilon_i}{D_i},\nonumber
\end{align}
where (a) is from the suboptimality of $N_i$.
Sequences $\{\gamma_i\}$, $\{Q_i\}$, and $\{D_i\}$ generated by FQL Algorithm satisfy
$ \gamma_i-\gamma^*\geq \gamma_i-{N_i}/{D_{k}}=-{Q_i}/{D_i}\geq  {\epsilon_i}/{\alpha D_i} $
for all $i$ such that $Q_i<0$. In addition, from the fact that
    $N_i-\gamma_i D_i\leq  N(\gamma^*)-\gamma_i D(\gamma^*)$ and 
    $N(\gamma^*)-\gamma^* D(\gamma^*)\leq  N_i-\gamma^* D_i$
for all $i\in[E]$, it follows that $
   (\gamma_i-\gamma^*)(D(\gamma^*)-D_i)\leq 0,~\forall i\in[E],$
and hence $ \gamma_i\geq \gamma^*~~{\rm if~and~only~if}~D_i\geq D(\gamma^*).$
It follows from \eqref{Proof-Eq1} that, for all $i$,
\begin{align}
    \gamma_{i+1}-\gamma^*=F_i-\gamma^*\leq~~& (\gamma_i-\gamma^*)\left(1-\frac{D(\gamma^*)}{D_i}\right)+\frac{\epsilon_i}{D_i}\nonumber\\
    \overset{(b)}{\leq} ~~& (\gamma_i-\gamma^*)\left(1+\alpha-\frac{D(\gamma^*)}{D_i}\right).\nonumber
\end{align}
Note that $(b)$ involves induction. Specifically, if $\gamma_{i}< \gamma^*$, then $\gamma_{i+1}< \gamma^*$. Therefore, we have that, if $Q_{i}<0$ then $Q_{i+1}<0$, and hence that $\epsilon_i\leq -Q_i$ for all $i\in[E]$.
Since $\alpha\in(0,1)$, it follows that $(\gamma_{i+1}-\gamma^*)/(\gamma_{i}-\gamma^*)\in(0,1)$ for all large enough $i$, which implies that $\{\gamma_i\}$ converges linearly to $\gamma^*$.

\section*{Appendix B: Analysis of Time Complexity}
\begin{proposition}\label{P1}
Proposed FQL Algorithm satisfies the stopping condition $\epsilon_i<-\alpha Q_i(\bs{s}_0,\bs{a}_i)$ for some $\alpha\in(0,1)$, then
after
\begin{align}
    T_i=\left\lceil\frac{11.66 \log(2|\mathcal{Z}|/(E\zeta))}{\alpha^2}\right\rceil
\end{align}
steps of SQL, the uniform approximation error $\norm{Q^*_{\gamma_i}-Q_i}\leq \epsilon_i$ holds for all $i\in[E]$, with a probability of $1-\zeta$ for any $\zeta\in(0,1)$.
\end{proposition}
\noindent{\textit{Proof Sketch}:} Specifically, we can prove that the required $T_i$ is proportional to ${Q_i^2}/{\epsilon_i^2}$, and hence corresponding to a constant upper bound. $\hfill\square$

Proposition \ref{P1} shows that\deleting{, surprisingly,} the total steps needed $T_i$ does not increase in $i$, even though the stopping condition $\epsilon_i<- \alpha Q_i(\bs{s}_0,\bs{a}_i)$ is getting more restrictive as $i$ increases.

\section*{Appendix C: DRL Network}
\subsection{DDPG Module}
In DDPG, there are an \textit{actor} and a \textit{critic}.
An \textit{actor} is responsible for selecting an action under the current state. It consists of two neural networks: $\neta$ determines the action for the task updating scheme; $\netat$ determines an action for updating the critic. 
A \textit{critic} is used for evaluating the action selected by $\neta$. It contains two neural networks: $\netc$ computes a Q-value of the action selected by $\neta$ under the current state to evaluate the expected long-term cost of the selected action; $\netct$ computes a target-Q value, which is used for updating $\netc$.




\subsubsection{Action Selection} 
Let $\thetamu_m$ denote the parameter of {$\neta$} of mobile device $m\in\M$. 
When task $k-1$ has been processed, the task generator  observes state $\su_m(k)$ and  chooses an action:
\begin{equation}\label{update_act}
    \au_m(k)= \mu_m(\su_m(k)|\bs\theta^{\mu}_m)+\N_m,
\end{equation}
where $\N_m$ is an exploration noise, and $\mu_m(\su_m(k)|\thetamu_m)$ denotes the action policy of state $\su_m(k)$ with $\thetamu_m$. 

\subsubsection{Neural Network Training}


Let $\thetamum_m$, $\thetau_m$, and $\thetaum_m$ denote that parameter vector of $\netat$, $\netc$, and $\netct$, respectively. Upon processing task $k$, the DDPG module observes the cost $c_{m}(k)$ and stores experience $(\au_m(k), \su_m(k),c_{m}(k),\su_m(k+1))$ to the replay buffer. The DDPG module then randomly samples a set $\mathcal{K}_{b}$ of mini-batches to update the critic network $\thetau_m$ by minimizing the difference between the recent Q-value of the selected action under $\su_m(k)$ and a target Q-value $y_k$:
\begin{align}
    L=\frac{1}{|\mathcal{K}_{b}|}\sum_{k\in\mathcal{K}_{b}}\left(y_{m,k}- \qu_m(\su_m(k),\au_m(k)|\thetau_m) \right)^2,
\end{align}
and $y_{m,k}\!=\!c_{m}(k)\!+\!\delta \qu_m\!(\su_m(k\!+\! 1),\mu_m\!(\su_m(k+\!1)|\thetamum_m)\! \\ |\thetaum_m)$.
In addition, the DDPG module updates the actor policy $\thetamu_m$ using the sampled policy gradient: for all $m\in\mathcal{M}$,
\begin{align}
    \nabla_{\thetamu_m}J\approx \frac{1}{|\mathcal{K}_{b}|}\sum_{k\in\mathcal{K}_{b}}& \nabla_{\au_m} \qu_m (\su_m(k),\mu_m(\su_m(k)|\thetamu_m) |\thetau_m) \nonumber\\
    &\times \nabla_{\thetau_m} \mu_m(\su_m(k)|\thetamu_m).
\end{align}
Finally, the DDPG module uses soft  target updates, based on $
    \thetaum_m=\tau \thetau_m+(1-\tau) \thetaum_m$ and
     $\thetamum_m=\tau \thetamu_m + (1-\tau) \thetamum_m$, with a small $\tau$.

\subsection{D3QN Module}
The main idea is to learn a neural network that maps from each state in state space $\So$ to the Q-value of each action in discrete action space $\Ao$. 
After obtaining such a mapping, given any state, the scheduler of the mobile device can choose the action with the minimum Q-value to minimize the expected long-term cost. 
There are two neural networks: $\nete$ is used for action selection;  $\nett$ is used for computing a target Q-value, where this value approximates the expected long-term cost of an action under the given state. Both neural networks have the same neural network structure: a fully connected network with an advantage and value (A\&V) layer. The A\&V layer is responsible for learning the Q-value resulting from the action and state, respectively.

\subsubsection{Action Selection} 
Let $\thetao_m$ denote the parameter vector of  $\nete$. Let $\qo_m (\so_m(k),\a; \thetao_m)$ denote the Q-value function of action $\a$ under state $\so_m(k)$ with parameter vector $\thetao_m$. After task $k$ has been generated, the scheduler of mobile device $m\in\M$ observes state $\so_m(k)$ and chooses an action as follows
\begin{equation}\label{offload_act}
    \ao_m(k)=\left\{
    \begin{array}{ll}
      \text{a random action from $\Ao$}, & \text{w.p.} \quad\epsilon_r,\\
        \arg\min_{\bs{a}}\qo_m (\so_m(k),\bs{a}; \thetao_m), & \text{w.p.} \quad 1-\epsilon_r,
    \end{array} \right.
\end{equation}
where ‘w.p.’ refers to “with probability”, and $\epsilon_r$ is the probability of random exploration.

\subsubsection{Neural Network Training}
When task $k$ of mobile device $m$ has been processed,  D3QN module observes the cost $c_{m}(k)$ and stores experience $(\ao_m(k), \so_m(k), c_{m}(k), \so_m(k+1))$ to replay buffer.  Then, the D3QN module randomly samples mini-batches to update $\thetao_m$ by minimizing the difference between the recent Q-value of the selected action under the observed state and the target Q-value $\bs{\qt}_m=\{\qt_{m,k}\}_{k\in\mathcal{K}_b}$:
\begin{equation}\label{eq:loss}
L(\thetao_{m},\bs{\qt}_{m})  = \frac{1}{|\mathcal{K}_b|}\sum_{k\in\K_b}\Big(\qo_m (\so_m(k),\a_m(k); \thetao_m)
- \qt_{m,k} \Big)^{2}.
\end{equation}
The target Q-value is an approximation of the Q-value by considering the next state and action. Let $\thetaom_m$ denote the parameter vector of $\nett$. The target Q-value is computed with $\nett$: 
\begin{equation}\label{eq:q-target1}
\qt_{m,k} = c_{m}(k) + \delta \qo_{m}(\s_{m}(k+1),\Atmp_{m,k};\thetaom_{m}),
\end{equation}
where $\Atmp_i$ is the action that minimizes the Q-value under the next state $\s_{m}(k+1)$ with $\nete$, i.e., for all $k$ and $m$, 
\begin{equation}
\Atmp_{m,k} =\arg \min_{\a\in\Ao} \qo_{m}(\s_{m}(k+1),\a;\thetae_{m}^o).
\end{equation} 

\subsubsection{Hyperparameter of Neural Network}
For the D3QN networks, we use  RMSProp optimizer. The batch size is $32$, the learning rate is  $3\times 10^{-4}$, and the discount factor is $\delta=0.9$. The probability of random exploration $\epsilon_r$ in \eqref{offload_act}  is gradually decreasing from $1$ to $0.003$. For the DDPG networks, we use Adam as the optimizer. The batch size is $64$, and the learning rates are $1\times10^{-4}$ and $1\times10^{-3}$ for actor and critic networks, respectively. Detailed structures of above networks see the codes provided.

\section*{Appendix D: Additional Evaluation}
We evaluate the convergence and performance of our algorithms Frac. DRL (OFL) and Frac. DRL (OFL+U) under different hyperparameters and environment settings.

\subsection{Environment Setting}
The proposed DRL framework is trained online with an infinite-horizon continuous-time environment, where we train the D3QN and DDPG networks to upgrade the task updating and offloading decisions respectively with collected experience. We evaluate the convergence of variants of our proposed DRL framework Frac. DRL (OFL) and Frac. DRL (OFL+U) under various DRL network hyperparameters respectively. Basically, we consider 1000 episodes (1500 episodes if necessary) with constant time limit and update fractional coefficient $\gamma_{m,i}$ every 50 episodes and for every experiment point, we run three times and average the results in our evaluation. We develop our programme on AMD EPYC 7402 CPU and Nvidia RTX 3090Ti GPU with Utuntu 20.04. Detailed package versions are presented in our Code Appendix. We present default basic parameter settings of our MEC environment in Table \ref{tab:env_para}.

\begin{table}[htbp]
    \centering
    \caption{Environment parameter settings}
    \label{tab:env_para}
    \begin{tabular}{@{}cccccc@{}}
        \toprule
        Parameter & Value \\ \midrule
        Number of mobile devices & 20 \\
        Number of edge nodes & 2 \\
        Capacity of mobile devices &  2.5 GHz \\
        Capacity of edge nodes &  41.8 GHz \\
        Task size & 30 Mbits \\
        Task density &  0.297 gigacycles per Mbits \\
        Drop coefficient &  1.5 \\
        \bottomrule
    \end{tabular}
\end{table}

\subsection{Hyperparameter Evaluation}
First we evaluate the convergence of the variant Frac. DRL (OFL) in Figure \ref{fig:hyper_d3qn}, which has offloading decisions with D3QN networks only and we keep hyperparamters to be the same among same type of networks from different mobile devices. In the figures, the $x$-axis represents the training episode, and the $y$-axis shows the averaged AoI among the mobile devices in each episode. The performance evaluations are plotted under different hyperparameters of the neural networks and we denote the random scheduling policy ``Random", where we randomly choose the offloading actions. 

Each mobile device performs the proposed algorithm in a decentralized manner
without interacting with other mobile devices. Note that even under
such an independent training framework, the scheduling policy
of each mobile device can gradually improve and converge. This algorithm contains three modules: cost module, DDPG module, and D3QN module. The cost module determines the cost function for the DDPG and the D3QN modules
based on the proposed fractional RL framework. The DDPG and the D3QN modules are responsible for making the task updating and offloading decisions, respectively.

Apart from these decisions made by neural networks, we have an additional dropping scheme in our algorithm. When the task processing duration of a mobile device exceeds the limitation, the task is dropped and start a new task immediately meanwhile the recorded AoI of the mobile device keeps increasing. This scheme can significantly better the performance.

\begin{figure}[t]
\centering
\includegraphics[height=3.29cm]{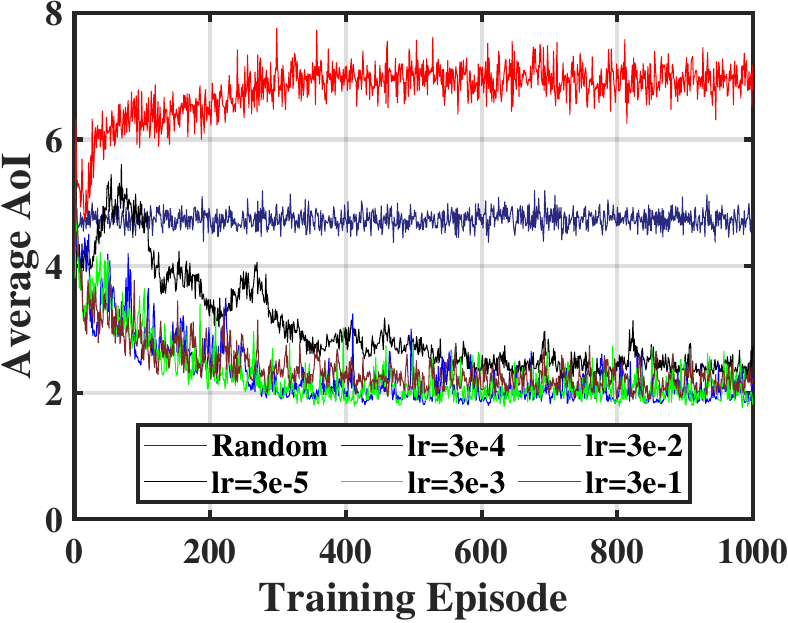}\label{fig:d3qn_lr}
\includegraphics[height=3.23cm]{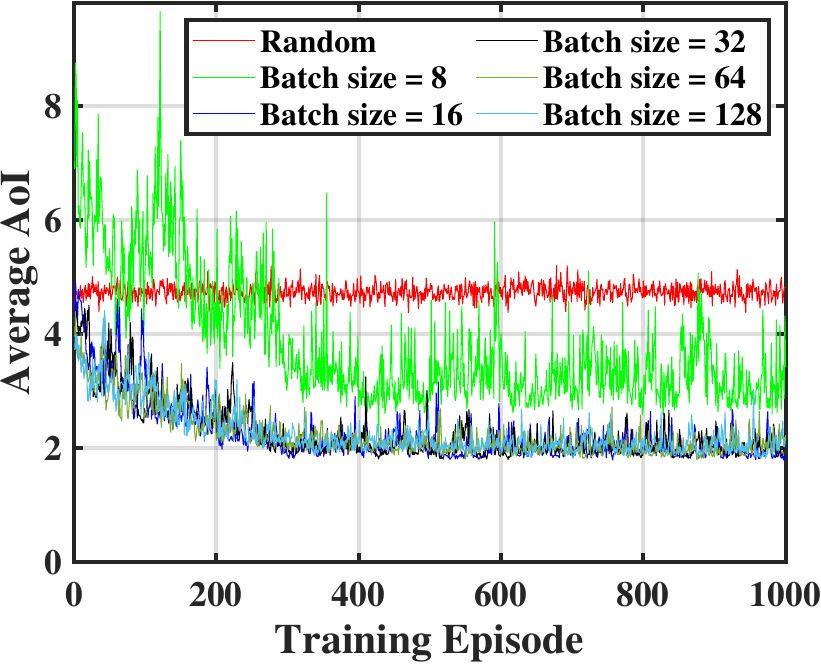}\label{fig:d3qn_batch}\\
\quad(a)\qquad\qquad\qquad\qquad\qquad\qquad(b)
\includegraphics[height=3.3cm]{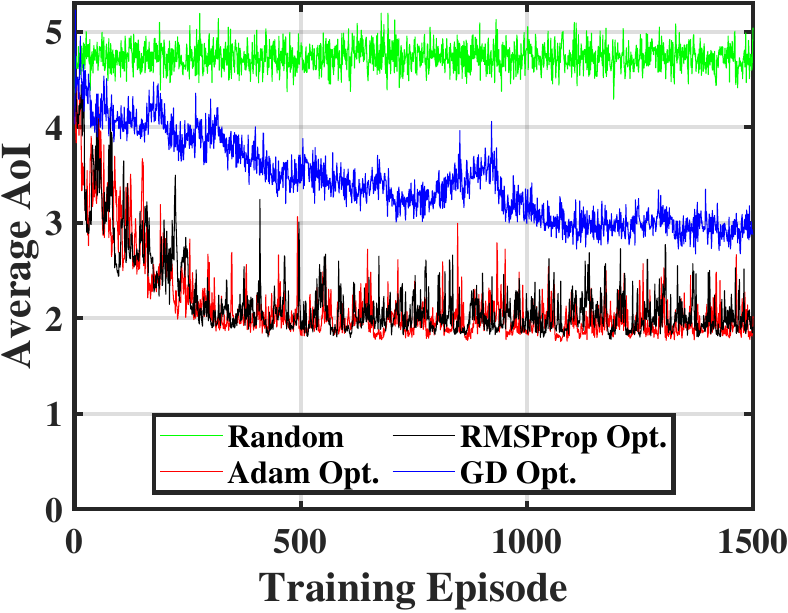}\label{fig:d3qn_opt}\\
(c)
\caption{
Frac. DRL (OFL) convergence and performance under different  (a) learning rates,
(b) batch sizes, (c) optimizers.}
\label{fig:hyper_d3qn}
\end{figure}

Fig.\ref{fig:hyper_d3qn}(a) shows the convergence under different values of learning rates (denoted ``lr") of D3QN networks, which is used to scale the magnitude of parameter updates during gradient descent. As shown in Fig.\ref{fig:hyper_d3qn}(a),  lr = $3\times 10^{-4}$ leads to a relatively smooth convergence and small average AoI. If the learning rate is too large (i.e., $3\times 10^{-1}$), it will be hard to converge and if it is too small (i.e., $3\times 10^{-5}$), it converges slowly. Fig.\ref{fig:hyper_d3qn}(b) shows Frac. DRL (OFL) performance under different batch sizes, i.e., the number of samples that will be propagated through the network. We can see when batch size is 32, the algorithm results in a promising performance and the result gets worse when batch size is too small (i.e. batch size = 8). Fig.\ref{fig:hyper_d3qn}(c) shows Frac. DRL (OFL) performance under different optimizers, which consist of adaptive moment estimation (Adam), gradient descent (denoted by "GD") and RMSprop optimizers, which are methods used to update the neural network to reduce the losses. In Fig.\ref{fig:hyper_d3qn}(c), RMSProp and Adam optimizers achieve similar convergent average AoI which is far better than gradient descent. 

\begin{figure}[t]
\centering
\includegraphics[height=3.23cm]{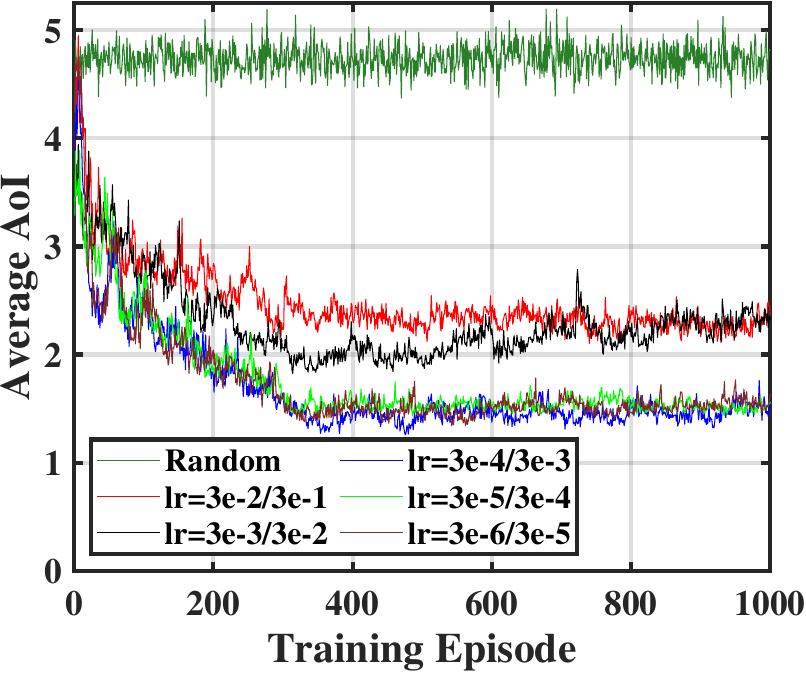}\label{fig:ddpg_lr}
\includegraphics[height=3.25cm]{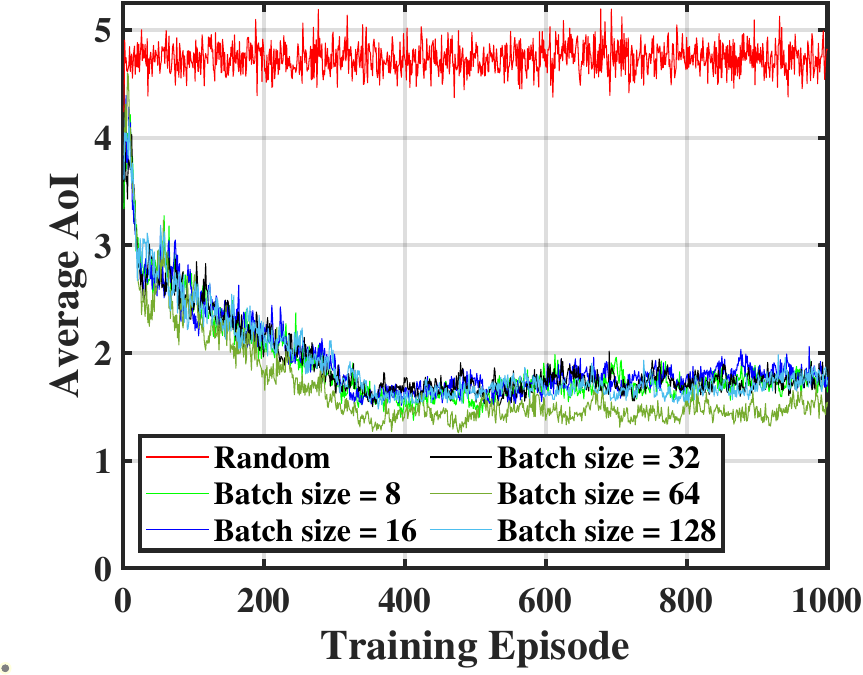}\label{fig:ddpg_batch}\\
\quad(a)\qquad\qquad\qquad\qquad\qquad\qquad(b)
\includegraphics[height=3.24cm]{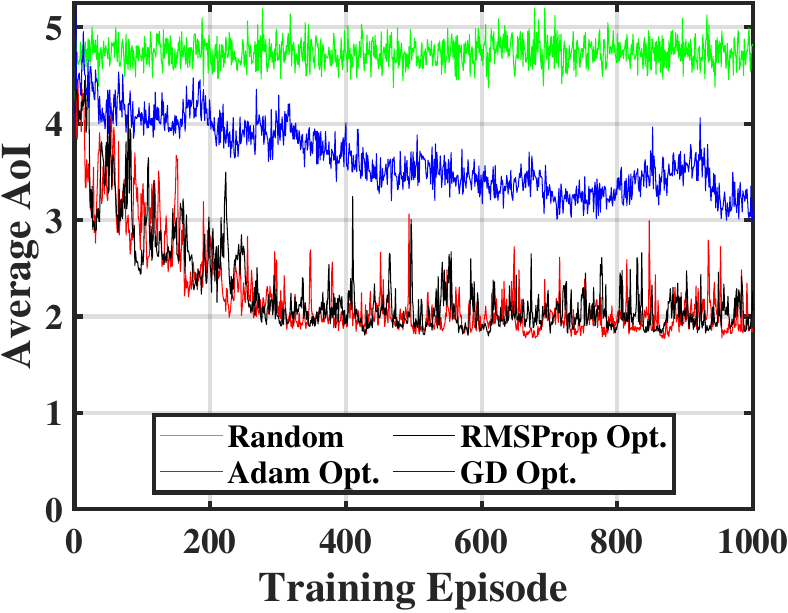}\label{fig:ddpg_opt}\\
(c)
\caption{
Frac. DRL (OFL+U) convergence and performance under different  (a) learning rates of Actor and Critic Networks respectively,
(b) batch sizes, (c) optimizers.}
\label{fig:hyper_ddpg}
\end{figure}

Then, we keep the hyperparameters of D3QN networks be constant and experiment the convergence and performance of Frac. DRL (OFL+U) algorithm under different hyperparameters of DDPG networks which are kept the same among different mobile devices. Fig.\ref{fig:hyper_ddpg}(a) shows the convergence under different values of learning rates of DDPG networks. As shown in Fig.\ref{fig:hyper_d3qn}(a), the learning rates of Actor and Critic being $3\times 10^{-4}$ and $3\times 10^{-3}$ respectively leads to a relatively fast convergence and small convergent average AoI. If the learning rates get larger, it results in worse convergence and performance. Fig.\ref{fig:hyper_ddpg}(b) shows Frac. DRL (OFL) performance under different batch sizes. We can see the convergent average AoI when batch size is 64 is better than the performance when batch sizes increase or decrease. Fig.\ref{fig:hyper_d3qn}(c) shows Frac. DRL (OFL) performance under different optimizers, where all three optimizers have similar performance. 

\begin{figure}[t]
\centering
\includegraphics[height=3.1cm]{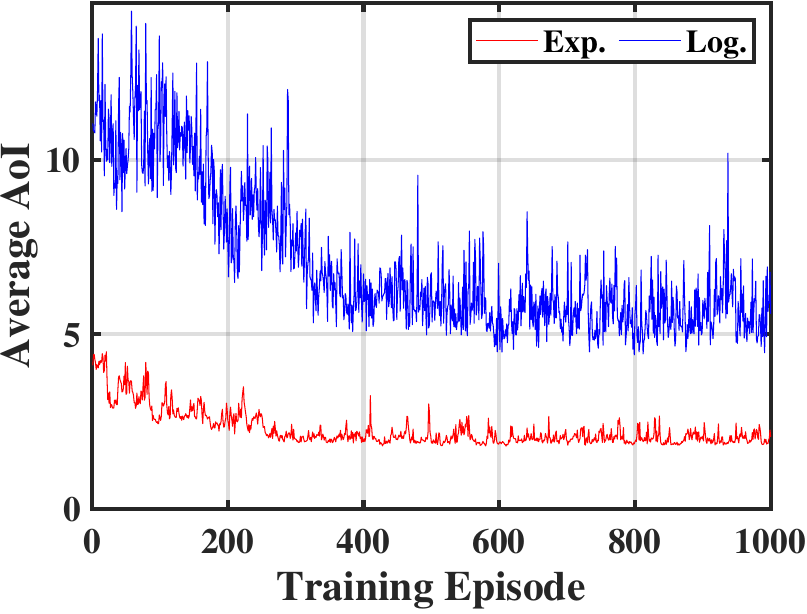}
\includegraphics[height=3.1cm]{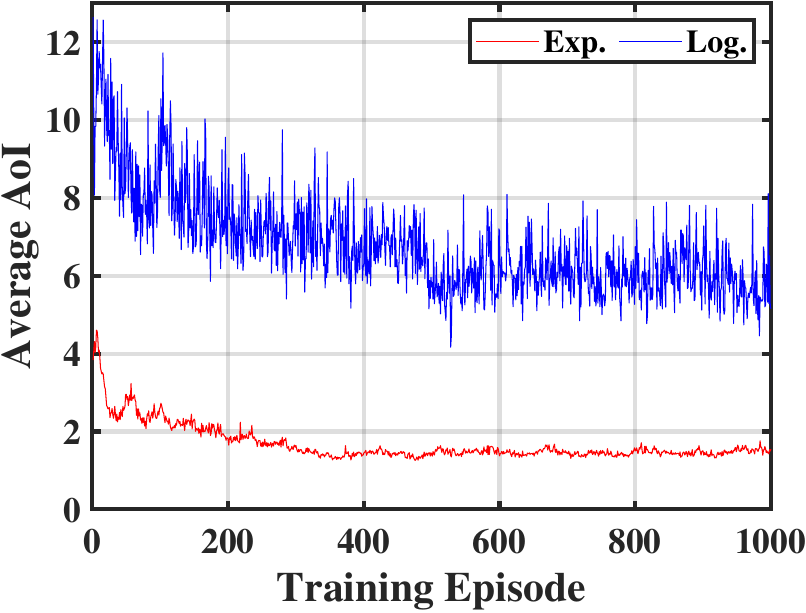}\\
\quad(a)\qquad\qquad\qquad\qquad\qquad\qquad(b)

\caption{Convergence under exponential and lognormal distributions of (a) of Frac. DRL (OFL), (b) Frac. DRL (OFL+U).}
\label{fig:num_mobile}
\end{figure}

\begin{figure}[H]
\centering
\includegraphics[height=3.3cm]{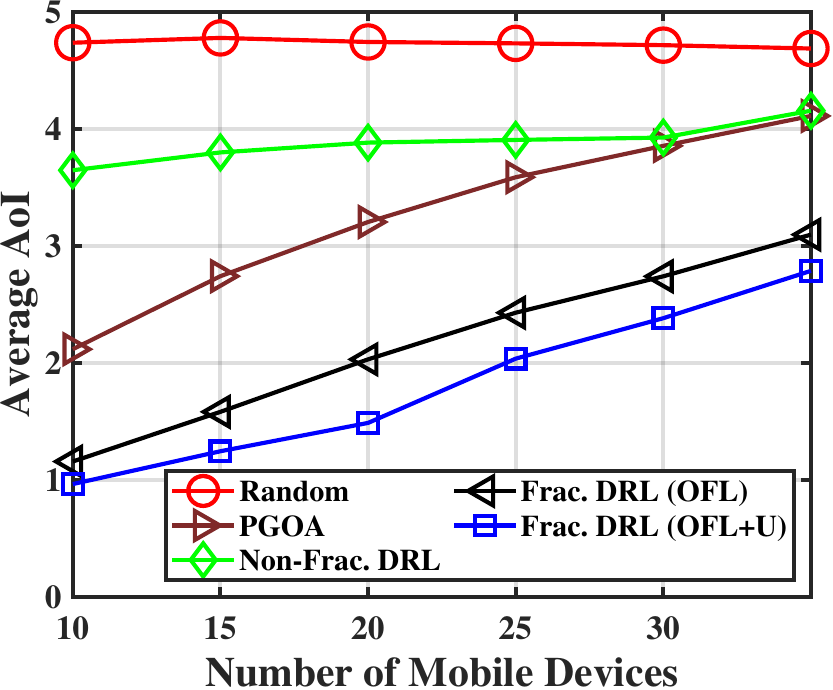}
\caption{
Performance comparison under different numbers of mobile devices.}
\label{fig:distribution}
\end{figure}

\subsection{Distributions of Duration} We experiment our algorithms under widely used exponential distribution and lognormal distribution, which is suitable for communication latency. In Fig.\ref{fig:distribution}, we show the convergence of our proposed algorithms Frac. DRL (OFL) and Frac. DRL (OFL+U) under these two distributions, where the exponential and lognormal distributions are denoted by "Exp." and "Log" respectively. It shows that both our proposed algorithms can converge with similar speed under these two distributions.

\subsection{Number of Mobile Devices} In Fig.\ref{fig:num_mobile}, we compare the algorithm
performance under different numbers of mobile devices. The performance gaps between the proposed fractional approaches and the non-fractional benchmarks are larger when the number mobile device is small. When the number of mobile devices is equal to 10, the Frac. DRL
(OFL+U) reduces the average AoI by 68.9\% when compared with Non-Frac. DRL. 

\end{document}